\documentclass[letterpaper]{article}
\usepackage{aaai2026}

\makeatletter
\def\copyright@text{}
\def\copyright@year{}
\gdef\copyright@on{}
\def\@copyright{}
\makeatother

\usepackage{times}
\usepackage{helvet}
\usepackage{courier}
\usepackage[hyphens]{url}
\usepackage{graphicx}
\usepackage{natbib}
\usepackage{caption}
\usepackage{algorithm}
\usepackage{algorithmic}

\usepackage{newfloat}
\usepackage{listings}
\DeclareCaptionStyle{ruled}{labelfont=normalfont,labelsep=colon,strut=off}
\floatstyle{ruled}
\newfloat{listing}{tb}{lst}{}
\floatname{listing}{Listing}

\usepackage{amsmath}
\usepackage[utf8]{inputenc}

\usepackage{url}
\usepackage{booktabs}
\usepackage{amsfonts}
\usepackage{nicefrac}
\usepackage{microtype}

\usepackage{tabularray}
\usepackage{pifont}

\usepackage{makecell}
\usepackage[table]{xcolor}
\usepackage{caption}
\usepackage{graphicx}
\usepackage{multirow}
\usepackage{colortbl}
\definecolor{customblue}{HTML}{74AED4}
\definecolor{customgreen}{HTML}{D3E2B7}
\definecolor{customred}{HTML}{ECA8A9}
\definecolor{custompurple}{HTML}{CFAFD4}
\definecolor{customorange}{HTML}{F7C97E}
\definecolor{deepgreen}{rgb}{0.0, 0.5, 0.0}

\definecolor{citecolor}{HTML}{0071BC}
\definecolor{linkcolor}{HTML}{ED1C24}

\usepackage[most]{tcolorbox}
\usepackage{xspace}
\tcbset{
  aibox/.style={
    width=\textwidth,
    top=10pt,
    colback=white,
    colframe=black,
    colbacktitle=black,
    enhanced,
    center,
    attach boxed title to top left={yshift=-0.1in,xshift=0.15in},
    boxed title style={boxrule=0pt,colframe=white,},
  }
}
\newtcolorbox{AIbox}[2][]{aibox,title=#2,#1}

\usepackage{tikz}

\newcommand{\squishlist}{
\begin{list}{$\bullet$}
{   \setlength{\itemsep}{0pt}
   \setlength{\parsep}{3pt}
   \setlength{\topsep}{3pt}
   \setlength{\partopsep}{0pt}
   \setlength{\leftmargin}{1.5em}
   \setlength{\labelwidth}{1em}
   \setlength{\labelsep}{0.5em} } }
\newcounter{Lcount}
\newcommand{\squishlisttwo}{
\begin{list}{\arabic{Lcount}. }
  { \usecounter{Lcount}
 \setlength{\itemsep}{0pt}
 \setlength{\parsep}{0pt}
 \setlength{\topsep}{0pt}
 \setlength{\partopsep}{0pt}
 \setlength{\leftmargin}{2em}
 \setlength{\labelwidth}{1.5em}
 \setlength{\labelsep}{0.5em} } }
\newcommand{\squishend}{\end{list} }

\newtcolorbox[list inside=prompt,auto counter,number within=section]{prompt}[1][]{
    colbacktitle=black!60,
    coltitle=white,
    fontupper=\footnotesize,
    boxsep=5pt,
    enhanced,
    left=0pt,
    right=0pt,
    top=0pt,
    bottom=0pt,
    boxrule=1pt,
    breakable,
    #1
}

\title{ShizhenGPT: Towards Multimodal LLMs for Traditional Chinese Medicine}
\author{
    Junying Chen\textsuperscript{\rm 1}$^\dagger$, Zhenyang Cai\textsuperscript{\rm 1}$^\dagger$, Zhiheng Liu\textsuperscript{\rm 1}, Yunjin Yang\textsuperscript{\rm 1}, Rongsheng Wang\textsuperscript{\rm 1}, \\
    Qingying Xiao\textsuperscript{\rm 3}, Xiangyi Feng\textsuperscript{\rm 2}, Zhan Su\textsuperscript{\rm 1}, Jing Guo\textsuperscript{\rm 5}, \\ Xiang Wan\textsuperscript{\rm 4}, Guangjun Yu\textsuperscript{\rm 1,\rm 3}, Haizhou Li\textsuperscript{\rm 1}, Benyou Wang\textsuperscript{\rm 1,\rm 4}$^*$\\
}
\affiliations{
    \textsuperscript{\rm 1} The Chinese University of Hong Kong, Shenzhen
    \textsuperscript{\rm 2} Freedom AI \\
    \textsuperscript{\rm 3} National Health Data Institute, Shenzhen
    \textsuperscript{\rm 4} Shenzhen Research Institute of Big Data \\
    \textsuperscript{\rm 5} Guangdong University of Technology \\
     \textit{wangbenyou@cuhk.edu.cn} \\
    \url{https://github.com/FreedomIntelligence/ShizhenGPT}
}

\usepackage{bibentry}

\begin{document}

\maketitle

\renewcommand{\thefootnote}{\fnsymbol{footnote}}
\footnotetext[2]{Equal Contribution. $^*$Corresponding author.}
\renewcommand{\thefootnote}{\arabic{footnote}}

\begin{abstract}
Despite the success of large language models (LLMs) in various domains, their potential in \textbf{Traditional Chinese Medicine (TCM)} remains largely underexplored due to two critical barriers: (1)  the scarcity of high-quality TCM data and (2) the inherently multimodal nature of TCM diagnostics, which involve looking, listening, smelling, and pulse-taking. These sensory-rich modalities are beyond the scope of conventional LLMs. To address these challenges, we present \textbf{ShizhenGPT}, the first multimodal LLM tailored for TCM. To overcome data scarcity, we curate the largest TCM dataset to date, comprising  100GB+ of text and 200GB+ of multimodal data, including 1.2M images, 200 hours of audio, and physiological signals.
 ShizhenGPT is pretrained and instruction-tuned to achieve deep TCM knowledge and multimodal reasoning. For evaluation, we collect recent national TCM qualification exams and build a visual benchmark for Medicinal Recognition and Visual Diagnosis.
 Experiments demonstrate that ShizhenGPT outperforms comparable-scale LLMs and competes with larger proprietary models. Moreover, it leads in TCM visual understanding among existing multimodal LLMs and demonstrates unified perception across modalities like sound, pulse, smell, and vision, paving the way toward holistic multimodal perception and diagnosis in TCM. Datasets, models, and code are publicly available. We hope this work will inspire further exploration in this field.
\end{abstract}

\section{Introduction}

\begin{figure}[th!]
    \centering
    \includegraphics[width=0.9\linewidth]{{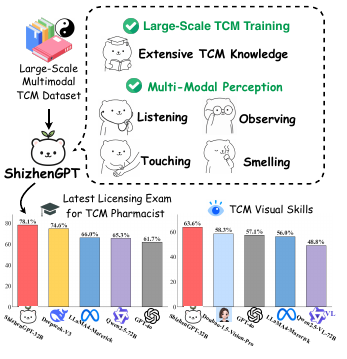}}
    \caption{Key Capabilities of ShizhenGPT, a multimodal LLM for Traditional Chinese Medicine (TCM).}
    \vspace{-4mm}
    \label{fig_head}
\end{figure}
Large language models (LLMs) like GPT-4 have demonstrated impressive capabilities across general and specialized domains \citep{gpt4,llama3,guo2025deepseek-r1}. Meanwhile, Traditional Chinese Medicine (TCM), a medical system with thousands of years of history and ongoing relevance for hundreds of millions~\cite{shi2007diagnosis,zhao2015advances}, remains largely absent from recent AI developments. TCM, with its intricate theoretical framework, nuanced diagnostic reasoning, and enduring clinical relevance~\cite{bing2010diagnostics}, presents both a challenge and an opportunity for large language models~\cite{lingdan}. Empowering these models to understand and reason about TCM could greatly advance clinical decision-making, medical education, and preservation of traditional medical knowledge~\cite{yao2019traditional,wang2020artificial}.

\begin{table*}[ht!]
\centering
\caption{Comparison of ShizhenGPT and existing LLMs across key dimensions.}
\resizebox{0.95\linewidth}{!}{
\begin{tabular}{lm{6cm}cccccm{3.5cm}}
\toprule
\multirow{2}{*}{$\mathbf{Model}$}& \multirow{2}{*}{\textbf{Domain Expertise}} & \multicolumn{5}{c}{\textbf{Multimodal Capability}} &  \multirow{2}{*}{\textbf{Clinical Application}} \\ 
\cline{3-7} & &  \makecell{\textbf{Text}\\ \includegraphics[width=0.17in]{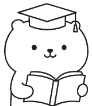}} & \makecell{\textbf{Vision}\\ \includegraphics[width=0.17in]{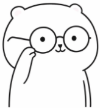}} & \makecell{\textbf{Sound}\\ \includegraphics[width=0.17in]{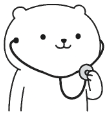}} & \makecell{\textbf{Smell}\\ \includegraphics[width=0.17in]{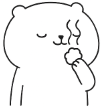}} & \makecell{\textbf{Pulse}\\\includegraphics[width=0.19in]{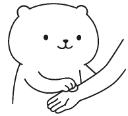}} & \\ 
\toprule
\textbf{TCM-specific LLMs}  &Limited by small-scale training data ({\textless}1GB). & \ding{51} & \ding{55} & \ding{55} & \ding{55} & \ding{55} & Limited to text-only scenarios. \\
 \midrule
 \makecell{\textbf{General-purpose LLMs}} & General-domain training, limited TCM specialization. & \ding{51} & \makecell{\ding{109}\\Partial} & \makecell{\ding{109}\\Partial} & \ding{55} & \ding{55} & Only partial support for vision/sound. \\
\midrule
\textbf{ShizhenGPT \includegraphics[width=0.16in]{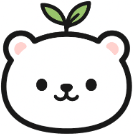}} & Trained on \textbf{15B+ tokens} and \textbf{200GB+} TCM data; high TCM proficiency. &   \ding{51} & \ding{51} & \ding{51} & \ding{51} & \ding{51} & Fully supports Four Diagnostic Methods.\\
\bottomrule
\end{tabular}
}
\label{tab:tcm_comparison}
\end{table*}

However, applying LLMs to TCM presents two key challenges.  First, there is a significant scarcity of high-quality TCM data. For example, most existing TCM-specific LLMs are trained on less than 1GB of text \cite{tcmchat, zhongjing, biancang, lingdan}. This is substantially smaller than the datasets used in domains like modern medicine \cite{zhang2024ultramedical, huatuogpt2}, and particularly insufficient given the complexity and depth of TCM theory. Moreover, the multimodal datasets necessary for training TCM-oriented multimodal LLMs are currently non-existent. Second, TCM fundamentally depends on multimodal diagnostics. Rooted in the "Four Diagnostic Methods", it depends on observing the tongue and other visual cues, listening to the voice and breath, smelling smells, and feeling the pulse~\cite{dong2022data,tian2023current}. These rich sensory modalities are central to TCM and lie well beyond the capabilities of text-only models.

To address this, we propose \textbf{ShizhenGPT}, the first multimodal LLM tailored for Traditional Chinese Medicine. To mitigate data scarcity, we built the largest TCM dataset to date, comprising over 100GB of text from 3,256 TCM-specific books and online sources. In addition, we collected over 200GB of multimodal data, including 1.2 million annotated images, 200+ hours of audio, and diverse physiological signals such as pulse, smell, and electrocardiograms (ECG). Leveraging this extensive dataset, ShizhenGPT is trained through domain-specific pretraining and instruction tuning, enabling it to acquire deep TCM knowledge and multimodal reasoning. The resulting model ShizhenGPT integrates deep domain expertise with perceptual capabilities aligned with real-world TCM diagnostic practices, including observation, listening, smelling, and pulse-taking.

To evaluate TCM-specific capabilities, we collected recent qualification exams, including three national licensing exams and two postgraduate entrance exams in TCM. All questions were released within the past year to ensure freshness. For visual evaluation, we built a benchmark using TCM atlases, containing 7k image-related questions focused on medicinal recognition and visual diagnosis. Results show that ShizhenGPT outperforms existing LLMs of similar scale and rivals much larger proprietary models, despite having only 32B parameters. In visual TCM tasks, it leads among current multimodal LLMs. Expert evaluations also indicate a higher preference for ShizhenGPT's responses. Additionally, across seven physiological signal datasets, ShizhenGPT demonstrates effective multimodal perception, including sound, pulse, and smell, achieving unified input and understanding across modalities.

We present a TCM-specific multimodal LLM capable of perceiving smell, sound, vision, and pulse. This expands the diagnostic capabilities of the model beyond text-based interaction, allowing it to directly observe the tongue, pulse, or breathing sounds for richer clinical insights. This moves toward more realistic, holistic medical AI systems. We hope our data, model, and approach will inspire further research in TCM.

Our contributions are as follows:

\squishlist

 \item We release the largest TCM dataset to date, covering extensive text and multimodal data across vision, audio, and physiological signals.

 \item We introduce ShizhenGPT, the first multimodal LLM tailored for TCM, capable of understanding images, sounds, smells, pulse, and more. It achieves strong performance in TCM expertise and leads in visual diagnostic tasks.

 \item We release a comprehensive benchmark for TCM across textual and multimodal tasks, enabling systematic evaluation of LLMs in TCM.

\squishend

\section{Background and Motivation}

\begin{figure*}[!ht]
    \centering
    \includegraphics[width=0.85\linewidth]{{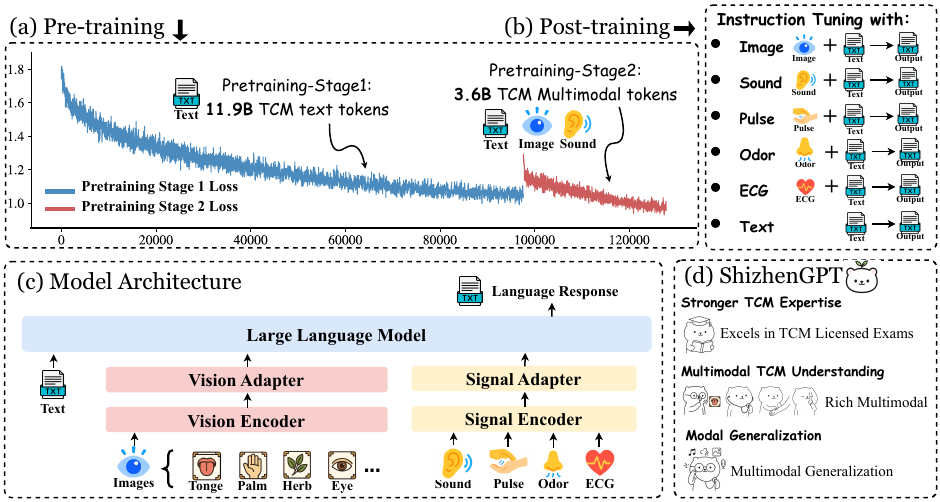}}
    \caption{Overview of ShizhenGPT. (a) Pre-training process, with the loss curve of \textit{ShizhenGPT-7B}. (b) Post-training process with multimodal instruction tuning. (c) Model architecture. (d) Demonstration of ShizhenGPT's capabilities.}
    \label{fig_method}
    \vspace{-2mm}
\end{figure*}

\paragraph{Traditional Chinese Medicine} TCM is one of the world's oldest and most influential medical systems, serving over 500 million patients annually in China and gaining global recognition \cite{cheung2011tcm,liu2015prevalence,zhao2023medical}. Unlike modern medicine, TCM relies on the "Four Diagnostic Methods": \textbf{looking}, \textbf{listening/smelling}, \textbf{questioning}, and \textbf{pulse-taking}~\cite{dong2022data,tian2023current}. These methods collect visual, auditory, verbal, and tactile information to form a holistic understanding of the patient. Each captures distinct biomedical signals, making TCM inherently multimodal and posing unique challenges for AI systems seeking to replicate its diagnostic reasoning.

\paragraph{Motivation for a  Multimodal TCM LLM} TCM's complexity and its reliance on diverse sensory modalities pose significant challenges for LLMs.
As shown in Table \ref{tab:tcm_comparison}, current TCM-specific models suffer from limited training data and lack multimodal perception. General-purpose models, while powerful, are not tailored to TCM and offer only partial support for modalities like vision or audio. To address these gaps, we propose \textbf{ShizhenGPT}, a multimodal LLM trained on extensive TCM data in diverse modalities. By integrating perception and reasoning across all four diagnostic methods, ShizhenGPT might enable more faithful, practical applications in real-world clinical settings.

\section{ShizhenGPT}

This section describes \textbf{ShizhenGPT}, including its architecture and two-phase training: (1) \textit{pre-training} to inject TCM-specific knowledge, and (2) \textit{post-training} for real-world tasks.

\subsection{Model Architecture}

As shown in Figure~\ref{fig_method}(c), ShizhenGPT consists of three main components: an LLM backbone, a vision encoder, and a signal encoder.

\paragraph{LLM Backbone}
The LLM serves as the core reasoning engine. It processes inputs from multiple modalities and generates responses. We use the \texttt{Qwen-2.5-7B} and \texttt{Qwen-2.5-32B} \cite{qwen} as the base LLMs.

\paragraph{Vision Encoder} For visual inputs, we initialize the \textit{Vision Encoder} from \texttt{Qwen-2.5-VL} \cite{Qwen2.5-VL}, which supports high-resolution images using 2D-RoPE and window attention. Visual patches extracted by the encoder are grouped (four adjacent patches), concatenated, and mapped to the LLM's embedding space via a two-layer MLP (\textit{Vision Adapter}).

\paragraph{Signal Encoder}
For continuous signals (e.g., voice, pulse, smell), we use \texttt{Whisper-large-v3} \cite{radford2022whisper} as the initial \textit{Signal Encoder}. Non-audio signals are first converted into waveform representations via linear interpolation.  All waveforms are resampled to 16kHz and transformed into 128-channel mel-spectrograms using a 25ms window and 10ms hop size.  A stride-2 pooling layer reduces temporal resolution, and a one-layer MLP (\textit{Signal Adapter}) projects the features into the embedding space of LLM. Although pulse signals are low-frequency, the resulting 40ms-per-token resolution retains sufficient temporal granularity to capture the slow-varying patterns essential for TCM.

\begin{table*}[!htbp]
\centering
\small
\caption{Overview of \textbf{Pre-training Datasets}. These data are utilized for a single training epoch unless otherwise specified. \textbf{\#} indicates quantity. \textbf{B} stands for billion. }
\resizebox{0.80\linewidth}{!}{
\renewcommand{\arraystretch}{1.2}
\begin{tabular}{lm{9.5cm}cc}
\toprule
\textbf{Pre-training Dataset} & \textbf{\qquad\qquad\qquad\qquad\qquad\qquad Description} & \textbf{\# Tokens} & \textbf{Storage}\rule{0pt}{2.8ex} \\
\hline
\rowcolor{cyan!10} \multicolumn{4}{c}{\includegraphics[width=0.16in]{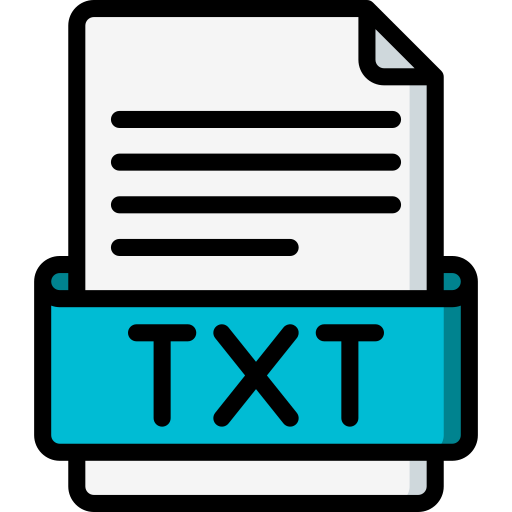} \textbf{Text Datasets (Pretraining Stage 1 \& Stage 2)}} \rule{0pt}{2.2ex}\\ \hline
\rowcolor{cyan!1} \includegraphics[width=0.16in]{images/txt-file.png} \textbf{TCM Web Corpus} & A high-quality TCM corpus collected from Common Crawl (2017–2023) and WeChat public articles (a major content channel in China). & \textbf{5.33B} & \textbf{21.2 GB} \\ \hline
\rowcolor{cyan!1} \includegraphics[width=0.16in]{images/txt-file.png} \textbf{TCM Book Corpus} & A cleaned corpus of 3,256 TCM textbooks. (\textbf{2 epochs}) & \textbf{0.96B}    & \textbf{3.8 GB} \\ \hline
\rowcolor{cyan!1} \includegraphics[width=0.16in]{images/txt-file.png} \textbf{General Text Corpus} & 	General-domain English corpus from FineWeb-edu. See Appendix~\ref{ap_general-corpus} for details. & \textbf{5.63B} & \textbf{29.2 GB} \\
\hline
\rowcolor{red!10}  \multicolumn{4}{c}{\includegraphics[width=0.16in]{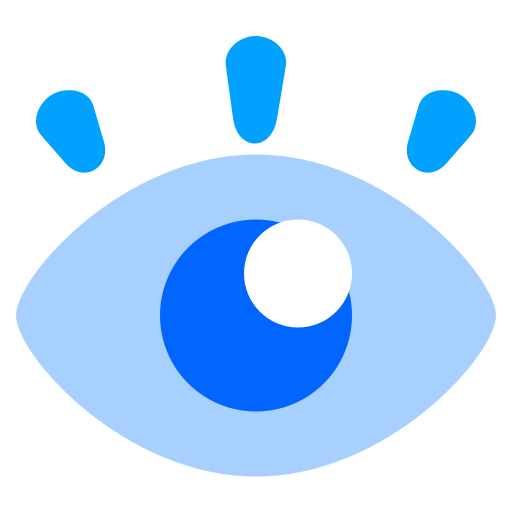} \textbf{Image-Text Dataset (Pretraining Stage 2)}}   \\ \hline
\rowcolor{red!2} \Gape[0pt][2pt]{\includegraphics[width=0.16in]{images/visual.png} \makecell[l]{\textbf{TCM Web} \\  \textbf{Interleaved Data}}}  & Interleaved text-image data from the TCM web corpus. (\# Images: \textbf{1,143,954}) & \textbf{0.87B} & \textbf{140.7 GB} \\ \hline
\rowcolor{red!2}  \Gape[0pt][2pt]{\includegraphics[width=0.16in]{images/visual.png} \makecell[l]{\textbf{TCM Book} \\ \textbf{Interleaved Data} }}
 & Richly illustrated interleaved text-image data from 306 TCM books.  (\# Images: \textbf{51,619}; \textbf{2 epochs}) & \textbf{0.14B} & \textbf{17.6 GB} \\ \hline
\rowcolor{red!2} \Gape[0pt][2pt]{\includegraphics[width=0.16in]{images/visual.png} \makecell[l]{\textbf{TCM Synthesized} \\ \textbf{Image-Text Data} }} & TCM image-text pairs generated from images and their context using multimodal LLMs. (\# Images: \textbf{146,635}; \textbf{2 epochs}) & \textbf{0.16B} & \textbf{40.6 GB} \\
\hline
\rowcolor{red!2} \Gape[0pt][2pt]{\includegraphics[width=0.16in]{images/visual.png} \makecell[l]{\textbf{General} \\  \textbf{Image-Text Data}}}  & General-domain image-text datasets, such as ShareGPT-4V. See Appendix~\ref{ap_general-corpus} for details. (\# Images: \textbf{1,851,321}) & \textbf{0.69B} & \textbf{93.29 GB} \\
\hline
\rowcolor{orange!12} \multicolumn{4}{c}{\includegraphics[width=0.16in]{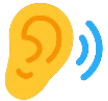} \textbf{Audio-Text Dataset (Pretraining Stage 2)}} \rule{0pt}{2.2ex} \\ \hline
\rowcolor{orange!2} \Gape[0pt][2pt]{\includegraphics[width=0.17in]{images/sound2.png} \makecell[l]{\textbf{TCM-related} \\  \textbf{Audio-Text Data}}}  & Chinese audio-text pairs in the domain of traditional Chinese medicine. (\# Audios: \textbf{58,462} ; Duration: \textbf{238.67 hours}) & \textbf{0.03B} & \textbf{75.72 GB} \\  \hline
\rowcolor{orange!2} \Gape[0pt][2pt]{\includegraphics[width=0.17in]{images/sound2.png} \makecell[l]{\textbf{General} \\  \textbf{Audio-Text Data}}}  & English audio-text pairs from general domains. See Appendix~\ref{ap_general-corpus}. (\# Audios: \textbf{282,884} ; Duration: \textbf{353.04 hours}) & \textbf{0.04B} & \textbf{111.6 GB} \\
\bottomrule
\end{tabular}
}
\label{tab:pretraining-datasets}
\end{table*}

\begin{table*}[h!]
\centering
\caption{Overview of \textbf{Instruction-Tuning Datasets}. Signal instructions are detailed in Appendix~\ref{ap_tcm_signal}.}
\resizebox{0.82\linewidth}{!}{
\renewcommand{\arraystretch}{1.2}
\begin{tabular}{lm{12.5cm}c}
\toprule
\textbf{Instruction-Tuning Dataset} & \textbf{\qquad\qquad\qquad\qquad\qquad Description} & \textbf{\# Item} \rule{0pt}{2.2ex} \\
\hline
\rowcolor{cyan!10} \multicolumn{3}{c}{\includegraphics[width=0.16in]{images/txt-file.png} 
\textbf{Text Instruction Data}} \\ \hline
\rowcolor{cyan!1}
\includegraphics[width=0.16in]{images/txt-file.png} \textbf{TCM Instruction}  & A dialogue instruction dataset centered on real-world TCM problems. & \textbf{83,629} \\
\hline
\rowcolor{red!10} \multicolumn{3}{c}{\includegraphics[width=0.16in]{images/visual.png} \textbf{Vision Instruction Data (Requires Vision Encoder)}} \\
\hline
\rowcolor{red!2}
\includegraphics[width=0.16in]{images/visual.png} \textbf{TCM Vision Instruction}  & A multimodal instruction dataset composed of TCM-related image-text pairs. \newline (\# Images: \textbf{70,638}) & \textbf{65,033} \\
\hline
\rowcolor{orange!12} \multicolumn{3}{c}{\includegraphics[width=0.16in]{images/sound2.png} \includegraphics[width=0.11in]{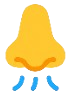} \includegraphics[width=0.18in]{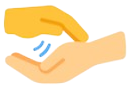} \includegraphics[width=0.16in]{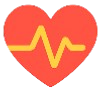} \textbf{Signal Instruction Data (Requires Signal Encoder)}} \\ \hline
\rowcolor{orange!2}
% TCM Speech Instruction Data & Speech-based instruction data in the TCM domain, possibly involving both speech and image inputs. (\# Audios: \textbf{57,957}; \# Images: \textbf{38,846}) & \textbf{57,957}\\
\includegraphics[width=0.16in]{images/sound2.png} \textbf{TCM Speech Instruction} &Speech-to-text instruction data in the TCM domain. & \textbf{57,957}\\ \hline
\rowcolor{orange!2}
\includegraphics[width=0.16in]{images/sound2.png} \textbf{Cough Sound Dataset} & An instruction dataset focused on cough sounds for medical diagnosis.  & \textbf{456} \\ \hline
\rowcolor{orange!2}
\includegraphics[width=0.16in]{images/sound2.png} \textbf{Heartbeat Sound Dataset} & An instruction dataset focused on heartbeat sounds for medical diagnosis.  & \textbf{189} \\ \hline
\rowcolor{orange!2}
\includegraphics[width=0.16in]{images/pulse2.png} \textbf{Pulse Dataset} & An instruction dataset focused on pulse diagnosis in TCM applications. & \textbf{4,101} \\ \hline
\rowcolor{orange!2}
\includegraphics[width=0.16in]{images/odour2.png} \textbf{Smell Dataset} & Dataset related to olfactory input for medical diagnosis. & \textbf{672} \\ \hline
\rowcolor{orange!2}
\includegraphics[width=0.16in]{images/ecg2.png} \textbf{ECG Dataset} & Electrocardiogram instruction dataset for medical diagnosis. & \textbf{99,195} \\ 
\bottomrule
\end{tabular}}
\label{tab:instruction-datasets}
\end{table*}

\subsection{Pre-training}

We begin with pre-training on large-scale TCM data to acquire domain expertise. We design a two-stage pipeline: the first stage infuses knowledge from extensive TCM text, while the second introduces multimodal alignment through image-text and audio-text data.

\paragraph{Pretraining Datasets}  
As summarized in Table~\ref{tab:pretraining-datasets},  the pretraining corpus  includes three components:

\squishlist

\item \textbf{TCM Text Corpus}  We collect TCM text from 3,256 TCM books and online sources. The books are converted into interleaved text-image corpora and cleaned to yield 3.8GB of \textit{book corpus}. For online sources, we use a 30K-term TCM lexicon to extract TCM documents from Common Crawl (2017-2023) and WeChat public articles, totaling 96.4GB. A two-step filtering process, involving quality scoring using a classifier and semantic deduplication, produces a final 21.2GB high-quality \textit{web corpus}.

\item \textbf{TCM Image-Text Data}
We extract 17.6GB of interleaved image-text data from TCM books (51K images). From 5 million images sourced from WeChat articles, we filtered high-quality TCM images using an image classifier. This process yielded 140.7GB of web-based image-text data (1M images). To improve alignment, we follow HuatuoGPT-Vision and use a multimodal LLM to generate descriptive captions based on image and context, producing 40.6GB of synthetic pairs (15K images).

\item \textbf{TCM Audio-Text Data}  
We extract 60K doctor-patient dialogues from the Huatuo-26M dataset  \cite{li2023huatuo26m} and synthesize audio via a high-fidelity TTS system \cite{ttsmodel}, yielding 58K utterance-aligned audio-text pairs.
\squishend

Further details are provided in Appendix~\ref{ap_pretraining_dataset}.

\paragraph{Pre-training Strategy}
We adopt a two-stage pre-training strategy. In \textbf{Stage 1}, it focuses on text-only learning with 11.9B tokens, including 6.3B from TCM corpora and 5.6B from general corpora (see Appendix~\ref{ap_general-corpus} for details) to maintain foundational capabilities.  In \textbf{Stage 2}, it uses 3.8B tokens of multimodal data, including TCM and general image-text and audio-text data, along with 1.8B resampled textual tokens from Stage 1 to preserve grounding. In both stages, inputs are packed into 4096-token sequences, and loss is computed only over textual tokens. Pre-training runs for 1 epoch with a 5e-5 learning rate using full-parameter tuning. Details are provided in Appendix \ref{ap_taining_detail}.

\subsection{Post-training}
After pre-training, the model acquires foundational TCM knowledge and multimodal abilities. Post-training aligns the model for instruction-following, extends its capabilities to downstream tasks, and enables adaptation to additional modalities such as sound and smell, which require low learning complexity but suffer from limited data.

\paragraph{Instruction Dataset}  
As summarized in Table~\ref{tab:instruction-datasets}, instruction tuning data spans four categories:

\begin{table*}[ht!]
\centering 
\caption{Results of LLMs on the latest TCM exams, including the 2024 Licensed TCM Pharmacist Exam, 2024 Licensed TCM Physician Exam, 2024 Licensed TCM Assistant Physician Exam, and two TCM Postgraduate Entrance Exams (2024-2025). \includegraphics[width=0.12in]{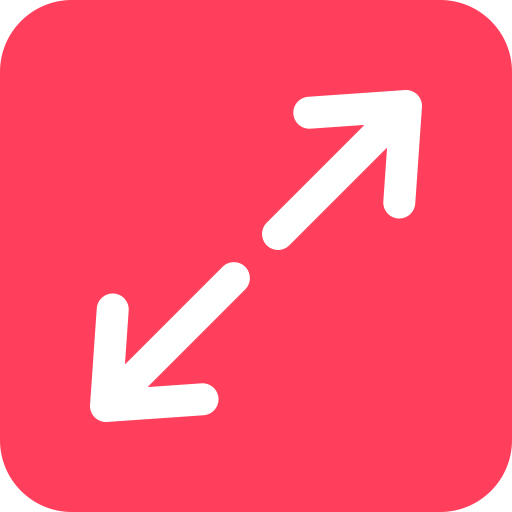} denotes proprietary or 100B+ LLMs; \includegraphics[width=0.12in]{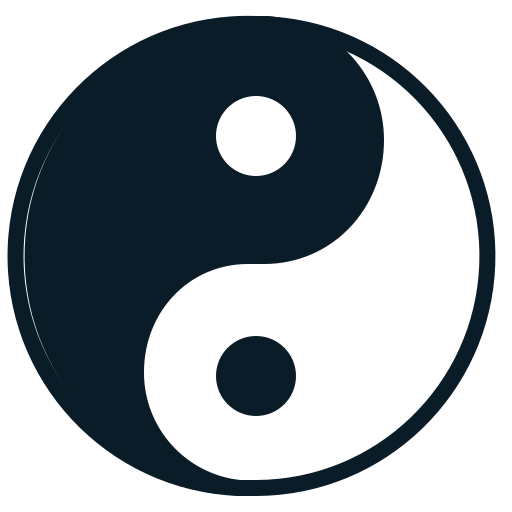} denotes TCM-specific LLMs.}
\resizebox{0.77\linewidth}{!}{
\renewcommand{\arraystretch}{1.0}
\begin{tabular}{lccccc|c}
\toprule
 $\mathbf{Model}$  & \makecell{ \footnotesize  $\mathbf{2024\ TCM}$  \\ \footnotesize$\mathbf{Pharmacist}$ } & \makecell{\footnotesize $\mathbf{2024\ TCM}$  \\ \footnotesize $\mathbf{Physician}$ } & \makecell{ \footnotesize $\mathbf{2024\ TCM}$  \\ \footnotesize$\mathbf{Assistant}$  \\ \footnotesize$\mathbf{Physician}$  }  & \makecell{ \footnotesize $\mathbf{2024\ TCM}$  \\ \footnotesize $\mathbf{Graduate}$ \\ \footnotesize$\mathbf{Entrance}$ }  & \makecell{ \footnotesize $\mathbf{2025\ TCM}$  \\ \footnotesize $\mathbf{Graduate}$ \\\footnotesize  $\mathbf{Entrance}$ }   & \makecell{   $\mathbf{Avg.}$}      \\ \midrule
 \multicolumn{7}{c}{\includegraphics[width=0.12in]{images/maximize.png} \textbf{Proprietary or 100B+ LLMs}} \\ \hline
\includegraphics[width=0.12in]{images/maximize.png} GPT-4o & 61.7 & 56.2  & 66.7 & 58.7 & 55.0 & 59.6   \\ 
\includegraphics[width=0.12in]{images/maximize.png}  LLaMA-4-Maverick-400B & 66.0 & 67.2  & 76.6 & 63.3 & 59.5 & 66.5   \\
\includegraphics[width=0.12in]{images/maximize.png} Deepseek-V3 & 74.6 & 74.8  & 80.8 & 72.4 & 66.7 & 73.9   \\
\includegraphics[width=0.12in]{images/maximize.png}  Doubao-1.5-Pro-32k & 81.9 & 79.7  & \textbf{86.2} & 81.5 & 73.7 & 80.6   \\
\includegraphics[width=0.12in]{images/maximize.png} Deepseek-R1-671B & \textbf{85.0} & \textbf{82.4}  & \textbf{86.2} & \textbf{83.2} & \textbf{83.3} & \textbf{84.0}   \\\midrule 
 \multicolumn{7}{c}{\textbf{{\textless}15B LLMs}} \\ \hline
%   \includegraphics[width=0.12in]{images/taichi.png} TCMChat-600k-7B & 22.0 & 29.5 & 26.0 & 21.6 & 18.5 & 23.5 \\
% \includegraphics[width=0.12in]{images/taichi.png} Zhongjing-7B & 25.5 & 27.8 & 26.4 & 18.4 & 25.6 & 24.7 \\
 % Gemma2-27B-it & 24.1 & 26.2  & 28.9 & 26.0 & 24.6 & 26.0   \\
 Llama-3.1-8B-Instruct & 36.7 & 36.9  & 40.4 & 33.4 & 26.1 & 34.7 \\
 \includegraphics[width=0.12in]{images/taichi.png} TCMChat-600k-7B & 42.0 & 49.5 & 46.0 & 41.6 & 38.5 & 43.5 \\
\includegraphics[width=0.12in]{images/taichi.png} Zhongjing-7B & 45.5 & 47.8 & 46.4 & 38.4 & 45.6 & 44.7 \\
 \includegraphics[width=0.12in]{images/taichi.png} Biancang-7B & 57.3 & 57.9 & 66.2 & 43.5 & 37.5 & 52.5 \\
 GLM4-9B-Chat & 54.6 & 56.2  & 61.6 & 54.0 & 51.9 & 55.7   \\
 Yi-1.5-9B-Chat & 59.0 & 58.2  & 69.7 & 53.0 & 39.7 & 55.9   \\
 % Qwen3-8B  & 58.3 & 59.9  & 73.3 & 57.7 & 52.3 & 60.3   \\
 \includegraphics[width=0.12in]{images/taichi.png} Lingdan-13B-PR & 64.4 & 65.9 & 78.4 & 45.6 & 41.1 & 59.1 \\
 DeepSeek-R1-Distill-14B & 62.2 & 60.8  & 70.5 & 58.4 & 55.7 & 61.5   \\
 Qwen2.5-7B-Instruct & 65.3 & 63.2  & 76.0 & 59.0 & 54.0 & 63.5   \\
 InternLM2.5-7B-chat & 69.8 & 68.5  & \textbf{79.7} & 58.5 & 48.9 & 65.1   \\
 Qwen2.5-14B-Instruct & \textbf{70.8} & 68.8  & 71.7 & 63.8 & 57.2 & 66.4   \\
\rowcolor[rgb]{0.87,0.94,1} \includegraphics[width=0.12in]{images/taichi.png} \textbf{ShizhenGPT-7B (Ours)} \rule{0pt}{2.1ex}     & 70.1 & \textbf{76.7}  & 79.4 & \textbf{68.8} & \textbf{69.3} & \textbf{72.9}   \\
\midrule \multicolumn{7}{c}{\textbf{15B to 100B LLMs}} \\  \hline
Gemma-3-27B-it  & 42.1 & 40.9  & 44.9 & 35.1 & 28.1 & 38.2   \\
Llama-3.1-70B-Instruct & 50.6 & 52.8  & 65.0 & 42.5 & 39.9 & 50.2   \\
Yi-1.5-34B-Chat & 67.6 & 62.2  & 75.5 & 61.8 & 50.7 & 63.6   \\
InternLM2.5-20B-Chat & 66.5 & 66.1  & 73.8 & 57.9 & 53.8 & 63.6   \\
GLM-4-32B & 67.9 & 65.6  & 80.9 & 64.9 & 62.6 & 68.4   \\
DeepSeek-R1-Distill-32B & 70.6 & 67.7  & 78.4 & 65.3 & 61.2 & 68.6   \\
Qwen2.5-32B-Instruct & 76.1 & 69.2  & 78.3 & 65.5 & 61.8  & 70.2   \\
Qwen2.5-72B-Instruct & 77.2 & 70.6  & 81.0 & 63.3 & 59.9 & 70.4   \\
\rowcolor[rgb]{0.87,0.94,1} \includegraphics[width=0.12in]{images/taichi.png} \textbf{ShizhenGPT-32B (Ours)} \rule{0pt}{2.1ex}   & \textbf{79.2} & \textbf{78.4}  & \textbf{84.9} & \textbf{77.5} & \textbf{70.7} & \textbf{78.1}   \\ 
\bottomrule
\end{tabular}
}
\label{tab:res_text}
\end{table*}

\squishlist
\item \textbf{TCM Text Instructions} ~
We curate 92K TCM questions from doctor-patient dialogues \cite{li2023huatuo26m} and medical verifiable questions \cite{chen2024huatuogpto1medicalcomplexreasoning}. Responses were distilled using Deepseek-R1 and validated with Deepseek-V3 based on reference answers or doctor responses, resulting in 83K reliable Chinese instruction pairs.

\item \textbf{Vision Instructions} ~
We select TCM-related images with detailed captions from books and web sources. GPT-4o is used to generate paired visual questions and answers based on image content and context, producing 65K high-quality vision instruction pairs.

\item \textbf{Audio Instructions}  
For speech audio, we convert 20K text questions into TTS speech and pair them with original answers. An additional 20K transcription-style instructions are created.  For TCM-specific audio (e.g., coughs, heart sounds), classification tasks are reformulated into instruction format (e.g., “Listen and determine if the patient has COVID-19.”).

\item \textbf{Physiological Signal Instructions}
Non-audio signals (e.g., pulse, smell, ECG) are converted to waveform format and tagged with modality-specific tokens (e.g., \texttt{<Smell>}, \texttt{<Pulse>}). The pregnancy-related pulse data is constructed in-house, while the other data are obtained from public datasets. All tasks are transformed into instruction format using a unified signal-processing pipeline to ensure extensibility.

\squishend

The instruction and signal datasets are detailed in Appendix~\ref{ap_instruction_dataset} and Appendix~\ref{ap_tcm_signal}, respectively.

\paragraph{Instruction Fine-tuning}  

We fine-tune the TCM-pretrained model on all modality-specific instruction datasets in Table~\ref{tab:instruction-datasets} using full-parameter tuning.  Instructions follow the \textit{Qwen-2.5} chat template, and loss is computed only on response tokens.  The model is trained for 3 epochs with a learning rate of 5e-6. See Appendix \ref{ap_taining_detail} for details. This stage produces the final model, \textbf{ShizhenGPT}.

\section{Experimental Setup}

\paragraph{Model Training}  

We train two versions of ShizhenGPT: \textbf{ShizhenGPT-7B} and \textbf{ShizhenGPT-32B}, based on \texttt{Qwen2.5-7B} and \texttt{Qwen2.5-32B} backbones, respectively. Both use the same vision and signal encoders, and are trained with identical settings across all stages. Experiments are conducted on two DGX nodes, each with 8 A100 GPUs.

\begin{table*}[!ht]
\centering
\caption{Results on the \textbf{TCM-Vision Benchmark}, evaluating visual understanding in TCM. \includegraphics[width=0.12in]{images/maximize.png} denotes proprietary or 100B+ multimodal LLMs. \textbf{bold} indicates the best score; \underline{underlines} marks the second-best.}
\resizebox{0.88\linewidth}{!}{
\renewcommand{\arraystretch}{1.0}
\begin{tabular}{lcccccccc|c}
\toprule
 \multirow{2}{*}{$\mathbf{Model}$} &\multicolumn{3}{c}{\includegraphics[width=0.16in]{images/visual.png} $\mathbf{TCM\ Medicinal\ Recognition}$}&
  \multicolumn{5}{c|}{\includegraphics[width=0.16in]{images/visual.png} $\mathbf{TCM\ Visual\ Diagnosis}$}&
  \multirow{2}{*}{$\mathbf{Avg}$}  \\   \cmidrule(lr){2-4}\cmidrule(lr){5-9}& \makecell{ \footnotesize $\mathbf{TCM}$  \\ $\mathbf{Patent}$ } & \makecell{ \footnotesize $\mathbf{TCM}$  \\\footnotesize $ \mathbf{Material}$ }  & \makecell{\footnotesize $\mathbf{TCM}$  \\ $\mathbf{Herb}$ }  & $\mathbf{Holism}$ & \footnotesize $\mathbf{Tongue}$ & $\mathbf{Palm}$  & $\mathbf{Tuina}$ &   $\mathbf{Eye}$    &          \\ \midrule
\multicolumn{10}{c}{\textbf{Multimodal LLMs}} \\  \hline
LLaVA-1.5-7B & 25.6 & 26.9 & 27.7 & 25.8 & 27.0 & 19.2 & 16.9 & 23.3 & 24.1 \\
LLaVA-1.5-13B & 21.6 & 34.8 & 26.4 & 31.9 & 33.5 & 29.2 & 21.4 & 18.3 & 27.1 \\
LLaVA-Med & 26.2 & 35.4 & 23.1 & 24.8 & 28.7 & 26.5 & 43.5 & 38.4 & 30.8 \\
DeepSeek-VL2 & 30.0 & 39.1 & 26.6 & 28.2 & 32.4 & 30.8 & 47.2 & 42.0 & 34.5 \\
MedGemma-4B-it & 30.2 & 40.6 & 28.0 & 46.7 & 49.9 & 41.5 & 38.3 & 38.1 & 39.2 \\
Gemma-3-27B-it & 32.6 & 40.1  & 31.8 & 45.8 & 48.8  & 46.2 & 38.1 & 45.3 & 41.1   \\
Llama-3.2-11B-Vision & 33.1 & 43.5  & 31.1 & 50.0 & 53.0  & 44.5 & 41.8 & 41.3 & 42.3   \\
Janus-Pro-7B & 32.5 & 45.1  & 33.0 & 45.7 & 48.6  & 45.3 & 45.8 & 50.8 & 43.4   \\
HuatuoGPT-Vision-7B & 38.4 & 53.0 & 36.2 & 48.9 & 52.6 & 44.9 & 50.4 & 49.0 & 46.7 \\
InternVL3-8B & 38.8 & 58.6  & 39.0 & 58.5 & 54.8  & 45.2 & 53.3 & 47.1 & 49.4 \\
Qwen2.5-VL-32B-Instruct & 39.6 & 60.3  & 46.0 & 61.2 & 56.4  & 54.8 & 51.4 & 49.2 & 52.4 \\
Kimi-VL-Instruct-16B & 47.1 & 64.6  & 46.4 & 60.5 & 53.0  & 52.3 & 51.9 & 51.9 & 53.5  \\
InternVL3-38B & 46.1 & 65.8  & 48.8 & 58.2 & 57.3  & 49.4 & 54.8 & 49.7 & 53.7 \\
Qwen2.5-Omni-7B & 43.8 & 61.8 & 51.2 & 60.8 & 54.5 & 53.5 & 59.5 & 52.0 & 54.6
 \\
\includegraphics[width=0.12in]{images/maximize.png}  LLaMA-4-Maverick-400B & 46.0 & 67.8  & 52.0 & 61.0 & 62.5  & 55.0 & 54.8 & 49.2 & 56.0   \\
\includegraphics[width=0.12in]{images/maximize.png} Gemini-1.5-Pro & \underline{49.8} & 73.5  & 54.0 & 56.7 & 57.2  & 48.8 & 59.7 & 50.7 & 56.3   \\
\includegraphics[width=0.12in]{images/maximize.png}  GPT-4o & 48.0 & 70.6  & 54.5 & 63.2 & 57.9  & 57.4 & 57.5 & 47.8 & 57.1   \\
\includegraphics[width=0.12in]{images/maximize.png}  Doubao-1.5-Vision-Pro & \textbf{52.0} & \textbf{79.3}  & \textbf{64.0} & 59.0 & 62.8  & 45.7 & 58.2 & 45.2 & 58.3 \\
\midrule
\multicolumn{10}{c}{\textbf{ShizhenGPT (Ours)}} \\\hline
\rowcolor[rgb]{0.87,0.94,1} \textbf{ShizhenGPT-7B} \rule{0pt}{2.1ex}   & 47.4 & 61.7  & 49.2 & \underline{70.6} & \underline{65.5}  & \textbf{63.7} & \underline{64.1} & \underline{58.3} & \underline{60.1}   \\ 
\rowcolor[rgb]{0.87,0.94,1} \textbf{ShizhenGPT-32B}  & 49.5 & \underline{74.2}  & \underline{54.7} & \textbf{71.5} & \textbf{66.5}  & 60.9 & \textbf{65.9} & \textbf{65.3} & \textbf{63.6}   \\ 
\bottomrule
\end{tabular}
}
\label{tab:res_vision}
\end{table*}

\paragraph{Baselines}  
We compare ShizhenGPT with following baselines: \textbf{(1) Vanilla LLMs}: We evaluate mainstream general-purpose models including LLaMA \cite{llama3}, Qwen, Yi \cite{young2024yi}, GLM-4 \cite{glm}, InternLM \cite{team2023internlm}, Gemma \cite{team2024gemma}, DeepSeek \cite{deepseekv3}, GPT-4o (2024-11-20) \cite{gpt4}, and Doubao \cite{doubao2024}. TCM-specific LLMs include TCMChat-600K-7B \cite{tcmchat}, Zhongjing-7B \cite{zhongjing}, Biancang-7B \cite{biancang}, and Lingdan-13B-PR \cite{lingdan}. \textbf{(2) Multimodal LLMs (MLLMs)}: For vision understanding, we include Gemini \cite{team2023gemini}, GPT-4o, LLaVA \cite{llava}, Qwen2.5-VL \cite{Qwen2.5-VL}, InternVL \cite{chen2024internvl}, Janus-Pro \cite{chen2025janus}, Kimi-VL \cite{team2025kimivl} and HuatuoGPT-Vision \cite{huatuogptvion}; for audio understanding, we consider Qwen2-Audio \cite{qwen2audio}, Llama-Omni \cite{fang2024llamaomni}, and Moshi \cite{defossez2024moshi}.

\paragraph{Benchmarks}
To evaluate the TCM capabilities, we construct a multimodal benchmark suite covering text, vision, signals, and human evaluation. \textbf{(1) Text Benchmarks:} We collect recent national TCM exams in China, including three licensing exams (for pharmacists, physicians, and assistant physicians) and two postgraduate entrance exams (2024–2025). 
\textbf{(2) Vision Benchmark:} We compile 7,204 multiple-choice questions from seven TCM atlases, explicitly excluded from training, covering seven subfields in medicinal recognition and visual diagnosis. See Appendix \ref{ap_benchmark} for details. \textbf{(3) Human Benchmark:} We collect 90 real-world TCM questions and evaluate model responses, with licensed TCM doctors selecting preferred answers (see Appendix \ref{ap_benchmark}). \textbf{(4) Signal Benchmarks:} We use the test set of the dataset for smell, auscultation, pulse, and ECG signals, along with a custom pregnancy detection dataset using pulse data.

\section{Results}

\paragraph{Evaluation of TCM Expertise} Table~\ref{tab:res_text} reports the performance of various models on the latest National TCM Licensing exams and the TCM postgraduate entrance exams.  For fair comparison, models are grouped by parameter size. In the small-scale category ({\textless}15B), \textbf{ShizhenGPT-7B} achieves the highest average score, outperforming \texttt{Qwen2.5-7B-Instruct} by +9.4 points despite sharing the same backbone.  It also surpasses larger models such as \texttt{Qwen2.5-14B-Instruct}, and leads all compared TCM-specific models in this range. In the mid-scale range (15-100B), \textbf{ShizhenGPT-32B} achieves the strongest results among open-source LLMs, with an average score of 78.1. This exceeds multiple 70B+ models. Although it slightly lags behind ultra-large proprietary models like \texttt{DeepSeek-R1} (671B) and \texttt{Doubao-1.5-Pro-32K}, it demonstrates highly competitive performance with significantly fewer parameters (78.1 vs. 84.0).

\begin{figure}[ht!]
    \centering
    \includegraphics[width=1\linewidth]{{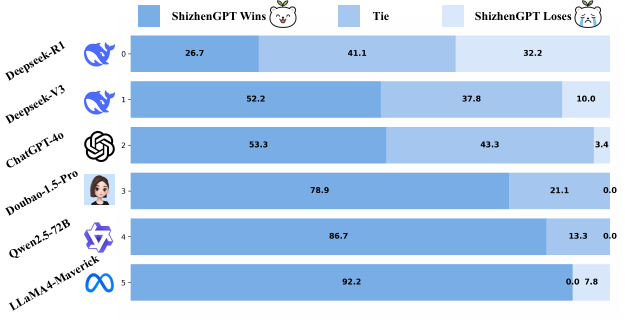}}
    \caption{Results of the human evaluation. ShizhenGPT refers to \textit{ShizhenGPT-32B}. "Win/Tie/Loss" indicates the proportion of expert preferences for the model responses (Details are in Appendix~\ref{app:humaneval}).}
    \label{tb_human_eval}
    \vspace{-4mm}
\end{figure}

\paragraph{Human Evaluation} To further assess TCM capabilities, we conduct an expert evaluation with licensed TCM practitioners. Following \cite{zhang2023huatuogpt}, we sample 90 real-world TCM queries. For each, our model and a baseline generate responses, which are then compared by experts. As shown in Figure~\ref{tb_human_eval}, ShizhenGPT-32B outperforms all baselines except DeepSeek-R1, including strong contenders like Doubao-1.5-Pro and DeepSeek-V3.  The gap with DeepSeek-R1 is small, while ShizhenGPT-32B offers a more efficient alternative given its much smaller size. Figure~\ref{tb_human_eval} reports the win/tie/loss results. ShizhenGPT-32B outperforms all baseline models except DeepSeek-R1, including strong contenders like Doubao-1.5-Pro and DeepSeek-V3. The performance gap with DeepSeek-R1 is minimal. Given the substantial inference cost of DeepSeek-R1 with 671B parameters, ShizhenGPT-32B offers a compelling trade-off between model size and clinical performance.

\paragraph{Evaluation of Visual Capabilities}  Table~\ref{tab:res_vision} presents results on the TCM visual benchmark, which evaluates the ability to recognize medicinal herbs and interpret diagnostic visuals (e.g., tongue, palm). Proprietary models like GPT-4o demonstrate stronger visual reasoning than open-source models such as Qwen2.5-VL. Among baselines, Doubao-1.5-Vision-Pro, a  proprietary MLLM, achieves the strongest baseline score of 58.3. \textbf{ShizhenGPT} surpasses all baselines, achieving a new SOTA with 63.6 using only 32B parameters. Notably, even the 7B version performs competitively, highlighting the effectiveness of domain-specific multimodal training for specialized visual understanding in TCM.

\begin{table}[!ht]
\centering
\caption{Results on general speech tasks.}
% \vspace{-1mm}
\resizebox{0.9\linewidth}{!}{
\begin{tabular}{lccc}
\toprule
\multirow{2}{*}{\textbf{Model}} & \multicolumn{3}{c}{\includegraphics[width=0.14in]{images/sound2.png} \textbf{General Audio LLMs Benchmarks}}\\ 
  & \textbf{LlamaQuestion} & \makecell{\textbf{Speech}\\\textbf{TriviaQA}} & \makecell{\textbf{Speech}\\\textbf{Web Questions}} \\
\midrule
Llama-Omni & 45.3 & 22.9 & 10.7  \\
Moshi & 43.7 & 23.8 & 16.7   \\
Qwen2-Audio & 60.0 & \textbf{30.4} &  24.0  \\
\rowcolor[rgb]{0.87,0.94,1}  \textbf{ShizhenGPT-7B} & \textbf{64.0} & 27.3 & \textbf{25.4}  \\
\bottomrule
\end{tabular}
}
\label{tab:speech}
\end{table}

\paragraph{Evaluation of General Audio Capabilities} Since ShizhenGPT is equipped with audio perception, we evaluate its general speech understanding ability on standard audio benchmarks. As shown in Table~\ref{tab:speech}, ShizhenGPT demonstrates strong performance, comparable to \textit{Qwen2-Audio-Instruction}. These results indicate that ShizhenGPT is not only compatible with various signal modalities but also maintains strong audio comprehension.

\begin{table}[!ht]
\centering
\caption{\label{tab:signal_res}Performance on various TCM signal modalities. All tasks are classification; details in Appendix \ref{ap_tcm_signal}. \textit{Random Baselines} indicate random prediction accuracy. }
% \vspace{-1mm}
\resizebox{1.0\linewidth}{!}{
\renewcommand{\arraystretch}{1.5}
\begin{tabular}{@{}clcc}
\toprule
\textbf{Modality} & \textbf{Task} & \makecell{\textbf{Random}  \\ \textbf{Baseline}}   & \cellcolor[rgb]{0.87,0.94,1}\makecell{\textbf{ShizhenGPT}  \\ \textbf{(7B)}} \\ \midrule
\multirow{2}{*}{\includegraphics[width=0.18in]{images/pulse2.png} $\mathbf{Pulse}$} 
    & Pregnancy Detection via Pulse Diagnosis & 50.0  & \cellcolor[rgb]{0.87,0.94,1}80.5 \\
    & Disease Classification via Pulse Diagnosis & 11.1 & \cellcolor[rgb]{0.87,0.94,1}20.6 \\
\hline
\includegraphics[width=0.12in]{images/odour2.png} $\mathbf{Small}$
    & Disease Classification via Smell Analysis & 14.3 & \cellcolor[rgb]{0.87,0.94,1}48.8 \\
\hline
\multirow{2}{*}{\includegraphics[width=0.16in]{images/sound2.png} $\mathbf{Sound}$} 
    & COVID-19 Detection from Cough Sound & 50.0 & \cellcolor[rgb]{0.87,0.94,1}58.7 \\
    & Cardiac Abnormality Detection from Heart Sound & 50.0 & \cellcolor[rgb]{0.87,0.94,1}62.9\\
\hline
\multirow{2}{*}{\includegraphics[width=0.16in]{images/ecg2.png} $\mathbf{ECG}$}
    & ECG Abnormality Detection & 50.0 & \cellcolor[rgb]{0.87,0.94,1}71.5\\
    & ECG Heart Beat Classification & 20.0 & \cellcolor[rgb]{0.87,0.94,1}83.1\\
\bottomrule
\end{tabular}
}
\vspace{-4mm}
\end{table}

\begin{table}[ht!]
\centering
\caption{Ablation study results on ShizhenGPT pretraining. Performance is evaluated across two TCM qualification exams and two TCM-specific visual benchmarks.}
\vspace{-2mm}
\resizebox{0.96\linewidth}{!}{
\begin{tabular}{lcccc}
\toprule
\multirow{2}{*}{\textbf{Model}} & \multicolumn{2}{c}{\textbf{TCM Expertise}} & \multicolumn{2}{c}{\textbf{TCM Visual Capability}} \\ 
  & \makecell{2024 TCM\\Pharmacist} &  \makecell{2024 TCM\\Physician} & \makecell{Medicinal\\Recognition} & \makecell{Visual\\Diagnosis} \\
\midrule
\rowcolor[rgb]{0.87,0.94,1}  ShizhenGPT-7B & 70.1\ \  & 76.4  & 52.7 & 64.5    \\
\quad\textbf{w/o} Pretraining & 66.9 (${\downarrow3.2}$) & 68.8 (${\downarrow7.6}$) & 43.7 (${\downarrow9.0}$) & 59.3 (${\downarrow5.2}$) \\ \midrule
\rowcolor[rgb]{0.87,0.94,1}  ShizhenGPT-32B & 79.2 & 78.4 & 59.5 & 66.0  \\
\quad\textbf{w/o} Pretraining & 76.9 (${\downarrow2.3}$) & 72.5 (${\downarrow5.9}$) & 51.3 (${\downarrow8.2}$) & 62.7 (${\downarrow3.3}$) \\
\bottomrule
\end{tabular}
}
\label{tab:ablation1}
\vspace{-2mm}
\end{table}

\paragraph{Evaluation of Signal Modalities} Table~\ref{tab:signal_res} shows model's performance across signal modalities, such as smell, ECG, and pulse. The model consistently outperforms random baselines, demonstrating its ability to unify diverse sensory inputs. Remarkably, it achieves 80\% accuracy in pregnancy detection from pulse signals alone. These findings provide empirical support for the diagnostic relevance of traditional sensory modalities in TCM.

\paragraph{Ablation Study on Pre-training}
We conduct an ablation study to evaluate the role of TCM-specific pre-training. As shown in Table~\ref{tab:ablation1}, it can be seen that extensive TCM pre-training substantially enhances the TCM expertise and visual understanding capabilities. This confirms that large-scale TCM training is effective for TCM-specific models.

\begin{figure}[!ht]
    \centering
    \includegraphics[width=0.85\linewidth]{{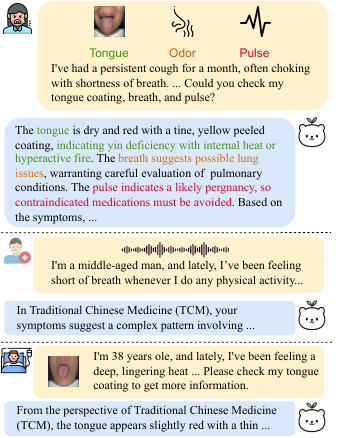}}
    \caption{Examples responses from ShizhenGPT-32B. Full outputs are provided in Appendix~\ref{app:case}.}
    \label{case_study}
\end{figure}

\paragraph{Case Study}
Figure~\ref{case_study} shows example responses from \textit{ShizhenGPT-32B} under multimodal inputs. The model can interpret user-provided photos, voice, and pulse signals, covering full TCM diagnostic scenarios and improving decision quality. These capabilities bring AI interaction closer to real-world clinical practice, highlighting ShizhenGPT's potential as a practical assistant in intelligent TCM diagnostics.

\section{Conclusion}
In this work, we introduce \textbf{ShizhenGPT}, the first multimodal LLM  tailored for TCM, along with an extensive TCM dataset. Experiments show that ShizhenGPT surpasses existing LLMs of a similar scale. It leads in visual TCM tasks and demonstrates strong multimodal perception, including sound, pulse, and smell. This extends its diagnostic ability beyond text, enabling direct analysis features such as the tongue, pulse, and breath sounds, for richer clinical insights. This work moves toward more holistic medical AI and aims to inspire further integration of TCM and AI.

\section*{Acknowledgments}

This work was supported by Shenzhen Medical Research Fund (No.C2406002) from the Shenzhen Medical Academy of Research and Translation (SMART), the Shenzhen Science and Technology Program (JCYJ20220818103001002), Shenzhen Doctoral Startup Funding (RCBS20221008093330065), Tianyuan Fund for Mathematics of National Natural Science Foundation of China (NSFC) (12326608), Shenzhen Science and Technology Program (Shenzhen Key Laboratory Grant No. ZDSYS20230626091302006), Shenzhen Stability Science Program 2023, and National Natural Science Foundation of China (NSFC) (72495131).

\bibliography{aaai2026}

\appendix
\clearpage

\section{Text Public Benchmark}

To further evaluate LLMs' TCM knowledge, we test the model on publicly available benchmarks, focusing on TCM-relevant subsets of MedQA (Chinese), CMB, CMExam, and CMMLU. As shown in Table~\ref{tab:res3}, ShizhenGPT achieves the best performance among open-source models under 100B parameters. The performance gap remains small, particularly in comparison with DeepSeek-R1, highlighting the importance of using real, up-to-date exam data for more meaningful evaluation.

\begin{table}[!ht]
\centering
\caption{Results on public benchmark datasets. $\ddag$ indicates the subset of questions related to traditional Chinese medicine (TCM).}
\resizebox{1.0\linewidth}{!}{
\begin{tabular}{lcccc}
\toprule
 $\mathbf{Model}$  &$\mathbf{MedQA}^\ddag$ & $\mathbf{CMB}^\ddag$ & $\mathbf{CMExam}^\ddag$  & $\mathbf{CMMLU}^\ddag$   \\ \midrule
Llama-3.1-70B-Instruct & 82.2 & 60.3  & 66.9 & 69.5   \\
Qwen2.5-7B-Instruct & 86.6 & 74.7  & 80.4 & 79.5   \\
internlm2\_5-20b-chat & 83.3 & 75.1  & 79.7 & 76.7   \\
Yi-1.5-34B-Chat & 86.1 & 77.3  & 81.7 & 78.9   \\
DeepSeek-R1-Distill-14B & 79.3 & 71.7  & 75.2 & 77.4   \\
Qwen2.5-72B-Instruct & 90.8 & 85.1  & 90.5 & 87.2   \\
Qwen2.5-32B-Instruct  & 89.9 & 84.8  & 88.7 & 86.9   \\
DeepSeek-R1-Distill-32B & 85.9 & 78.8  & 82.2 & 83.6   \\
GPT-4o & 84.0 & 70.7  & 76.7 & 75.0   \\
doubao-1-5-pro-32k-250115 & 93.7 & 91.0  & 94.8 & 90.0   \\
Deepseek-R1 & 93.7 & 92.2  & 93.5 & 90.6   \\
\midrule
\textbf{ShizhenGPT-7B}   & 90.0 & 79.0  & 85.3 & 84.4   \\
\textbf{ShizhenGPT-32B}  & 92.2 & 88.7  & 92.9 & 89.7   \\ 
\bottomrule
\end{tabular}
}
\label{tab:res3}
% \vspace{-4mm}
\end{table}

%%%%%%%%%%%%%%%%%%%%%%%%%%%%%%%%%%%%%%%%%%%%%%%%%%%%%%%%%%%%
\section{Related Work}

\paragraph{\textbf{Multimodal LLMs for Medical Specific Modalities}} Since the release of GPT-4~\cite{gpt4}, numerous powerful multimodal LLMs~\cite{llava,Qwen2.5-VL,team2023gemini,chen2024internvl} have emerged. To strengthen their capabilities in the medical domain, researchers have adapted these models to various medical modalities. As a core multimodal task, many MLLMs have been designed to interpret image-text inputs~\cite{awadalla2023openflamingo,li2023llavamed,wu2023towards,wu2024pmc,chen2024huatuogpt,tu2024towards,he2024meddr,su2025gmai}, supporting common medical images such as X-rays and pathology slides. Some even extend to 3D imaging like CT and MRI~\cite{wu2023generalist}, further expanding their applicability. Beyond imaging, medical modalities such as ECG signals are also being explored, with models now able to interpret them directly~\cite{zhao2024ecg} or process them as visual inputs~\cite{liu2024teach} to enhance understanding.

\paragraph{\textbf{LLMs in TCM}} LLMs have been fully adapted in medical domain~\cite{MedicalGPT, wang2023huatuo,han2023medalpaca, wu2024pmc, pal2024openbiollms, labrak2024biomistral, bao2023discmedllm,zhang2023biomedgpt,chen2024huatuogpt,wang2024apollo,apollomoe,christophe2024med42,singhal2023large}. The development of LLMs for TCM has made solid progress and yielded encouraging results. Some existing models~\cite{lingdan,haoyu2024tcmllm,yang2024tcm,liu2024cpmi,zhou2024tcm} have improved performance by collecting high-quality data on TCM prescription recommendations, syndrome differentiation, and QA, then applying instruction tuning to enable diagnosis from a TCM perspective. Other studies~\cite{tcmchat,biancang,tan2024medchatzh,zhang2024qibo,dong2024presrecst} have identified a lack of TCM knowledge in existing base models. Since fine-tuning alone could not fully address this, they compiled TCM resources from online forums and classical texts, incorporating this content through continued pretraining to enhance model performance. Besides that, models, like \cite{zhuang2025tcm} use RAG and knowledge graph to enhance LLMs' ability in TCM diagnosis. These efforts have laid an important foundation for the future of TCM LLMs.

\section{Additional Ablation Studies}

\begin{table*}[ht!]
\centering
\caption{Effect of pretraining data scale on downstream TCM tasks.}
\resizebox{0.8\linewidth}{!}{
\begin{tabular}{lcccc}
\toprule
\textbf{Model} & \makecell{\textbf{2024 TCM} \\ \textbf{Pharmacist Exam}} & \makecell{\textbf{2024 TCM} \\ \textbf{Physician Exam}} & \makecell{\textbf{TCM Visual} \\ \textbf{Medicinal Recognition}} & \makecell{\textbf{TCM Visual} \\ \textbf{Diagnosis}} \\
\midrule
\textbf{ShizhenGPT-7B} & & & & \\
\quad + 0\% Pretraining & 66.9 & 68.8 & 43.7 & 59.3 \\
\quad + 10\% Pretraining Data & 68.1 (\textcolor{blue}{+1.2}) & 70.2 (\textcolor{blue}{+1.4}) & 44.0 (\textcolor{blue}{+0.3}) & 61.4 (\textcolor{blue}{+2.1}) \\
\quad + 100\% Pretraining Data & \textbf{70.1} (\textcolor{blue}{+3.2}) & \textbf{76.4} (\textcolor{blue}{+7.6}) & \textbf{52.7} (\textcolor{blue}{+9.0}) & \textbf{64.5} (\textcolor{blue}{+5.2}) \\
\bottomrule
\end{tabular}
}
\label{tab:res1new}
\end{table*}

\subsection{Effect of Pretraining Data Scale}

To assess the impact of dataset size, we performed an ablation using different amounts of pretraining data. Due to time constraints, we report results with 10\% and 100\% of the full data compared to a model without pretraining.

As shown in Table~\ref{tab:res1new}, performance improves consistently with more pretraining data, highlighting the value of large-scale pretraining.

\begin{table*}[ht!]
\centering
\caption{Modality ablation results.}
\resizebox{0.8\linewidth}{!}{
\begin{tabular}{lccc}
\toprule
\textbf{Fine-tuning Setting} & \makecell{\textbf{Pulse Modality} \\ \textbf{Pregnancy Detection (\%)}} & \makecell{\textbf{Smell Modality} \\ \textbf{Disease Classification (\%)}} & \makecell{\textbf{Audio Modality} \\ \textbf{Cardiac Abnormality Detection (\%)}} \\
\midrule
Pulse Only & 79.5 & --- & --- \\
Smell Only & --- & 47.5 & --- \\
Audio Only & --- & --- & 63.4 \\
Pulse + Smell + Audio & \textbf{80.1} & \textbf{47.9} & 62.9 \\
\bottomrule
\end{tabular}
}
\label{tab:res3new}
\end{table*}

\subsection{Modality Contribution Analysis}

We also conducted an ablation to analyze the contribution of each modality (pulse, smell, and audio). Each modality was tested individually, as well as in a combined setting. As shown in Table~\ref{tab:res3new}, integrating multiple modalities yields modest gains in some tasks while preserving strong single-modality performance, demonstrating the flexibility of ShizhenGPT.

\section{General-domain Dataset}
\label{ap_general-corpus}
This section introduces the general-domain pre-training data incorporated during the pre-training phase to preserve the model's general capabilities, following \cite{zhang2024ultramedical,chen2024huatuogpto1medicalcomplexreasoning}. Due to the extensive scale of pre-training, prolonged domain-specific learning without revisiting general knowledge can lead to degradation of the LLM's foundational abilities. To address this, we included the following two categories of general-domain datasets.

\paragraph{General Text Dataset}
To maintain general language proficiency, we employed the FineWeb-edu corpus \cite{penedo2024finewebdatasetsdecantingweb}, a high-quality English pre-training dataset. A randomly sampled subset comprising 5.63 billion tokens was used in mixed pre-training to support the retention of general capabilities.

\paragraph{General Image-Text Dataset}
To enhance multimodal understanding during the second stage of pre-training, we incorporated general-purpose vision-language datasets. These included ShareGPT-4V \cite{chen2024sharegpt4v}, LLaVA Instruction Data \cite{llava}, ALLaVA-Caption-LAION-4V \cite{chen2024allava}, and the PubMedVision Alignment Dataset \cite{huatuogptvion}.

\paragraph{General Audio Dataset}
To facilitate alignment between speech and text, we introduced English general-domain audio-text pair datasets. Specifically, we utilized the dataset from AnyInstruct \cite{AnyInstruct}.

\section{Pre-training Dataset}
\label{ap_pretraining_dataset}

\begin{prompt}[title={The prompt for rating TCM text quality}] 
{
**Task Description:** \newline
Please evaluate the quality of the following paragraph related to Traditional Chinese Medicine (TCM) and assign a score (1–5) based on the criteria below: \newline

**Scoring Criteria:** \newline
**1 point**: The content is largely irrelevant to TCM or fails to provide any useful TCM knowledge. \newline
**2 points**: Contains minimal TCM content, or the TCM knowledge provided is extremely limited. \newline
**3 points**: Includes some TCM content, but the information is limited or contains advertisements or false/misleading claims (unreliable). \newline
**4 points**: Provides relatively reliable TCM knowledge and is suitable for model training. \newline
**5 points**: Offers rich and accurate TCM knowledge of high quality, suitable for high-quality training data. \newline

Please output a **number between 1 and 5** as the score only. Do not include any additional text. \newline

**Text to be evaluated:** \newline 
{\textless}TCM Paragraph{\textgreater} \newline
\textcolor{blue}{\texttt{\{TCM Paragraph\}}} \newline
{\textless}/TCM Paragraph{\textgreater} \newline
}
\end{prompt}
\captionof{figure}{\label{prompt-rating-text}The prompt for rating the quality of TCM documents. Here, \textcolor{blue}{\texttt{\{TCM Paragraph\}}} represents the TCM text to be scored.}

\paragraph{TCM Text Corpus}  
We collect TCM-related text from both classical books and online sources. For books, we gathered 5,000 TCM classical books and used the tool MinerU to convert scanned PDFs into interleaved text-image corpora. Non-informative content (e.g., covers, tables of contents) was removed, resulting in 3.8GB of clean book text. For online data, we constructed a 30K-term TCM lexicon to identify high-density TCM documents from Common Crawl (2017–2023) and WeChat public articles (primarily sourced from OmniCorpus-CW \cite{li2024omnicorpus}), yielding 33.2GB and 63.2GB of text corpora, respectively. To ensure quality, we applied a two-stage filtering pipeline: (1) we scored all documents using a classifier trained on DeepSeek-V3-labeled data with \texttt{text2vec-base-chinese} model \cite{text2vec}, assigning quality scores from 1 to 5 and removing low-quality or spammy content using the prompt in Figure \ref{prompt-rating-text}; (2) we performed semantic deduplication using cosine similarity on text embeddings to eliminate redundancy. The final high-quality TCM web corpus contains 21.2GB (5B tokens).

\paragraph{TCM Image-Text Corpus}
We extract 17.6GB of interleaved image-text data (51K images) from TCM books with rich illustrations (>50 images per book). Additionally, from 5 million WeChat articles containing images, we filtered TCM-relevant samples using an image classifier based on a CLIP vision encoder fine-tuned with GPT-4o-labeled data. This process yielded 140.7GB of web-based image-text data (~1M images). To enhance alignment quality, we followed HuatuoGPT-Vision and used Doubao-1.5-vision-pro (a multimodal LLM) to synthesize descriptive captions based on image and context using the prompt in Figure \ref{prompt-caption}, generating 40.6GB of synthetic high-quality image-text pairs.

\begin{prompt}[title={The prompt for generating descriptive captions for TCM images}] 
{
 \textcolor{blue}{\texttt{\{image\}}}\newline

Please generate a professional, detailed, and high-quality description for the medical image I provide. The description should include as many Traditional Chinese Medicine (TCM)-related visual details as possible to ensure clinical readability, professionalism, and comprehensiveness, while also being understandable to the general public. Aim to make the description as rich and detailed as possible, providing extensive visual information. \newline

You may refer to the image's context to improve the accuracy and completeness of your description, but do not reveal that you referenced the context. \newline

{\textless}caption{\textgreater}\newline
 \textcolor{blue}{\texttt{\{caption\}}}\newline
{\textless}/caption{\textgreater}\newline

{\textless}image context{\textgreater} ({\textless}image{\textgreater} represents the location of the image)\newline
 \textcolor{blue}{\texttt{\{image context\}}}\newline
{\textless}/image context{\textgreater}\newline

Please output a detailed description of the image only. Do not generate any content unrelated to the task.
}
\end{prompt}
\captionof{figure}{\label{prompt-caption}The prompt for generating questions for TCM Vision Instructions. Here, \textcolor{blue}{\texttt{{image}}} represents the TCM image. \textcolor{blue}{\texttt{{caption}}} and \textcolor{blue}{\texttt{{image context}}} represent the corresponding image caption and contextual information.}

\paragraph{TCM Audio-Text Corpus}  
We extracted 60K TCM-related doctor-patient dialogues from the Huatuo-26M \cite{li2023huatuo26m} dataset and synthesized corresponding audio using a high-fidelity TTS system \cite{ttsmodel}. This produced 58K audio-text pairs aligned at the utterance level, forming a TCM-specific audio-text dataset.

\section{Pre-Training Details}
\label{ap_taining_detail}

\paragraph{Stage 1: Text-Only Pre-training}  
Since textual data greatly outweighs multimodal data in token volume, we begin by pretraining the model on TCM-related textual corpora to establish strong domain knowledge. This stage uses 6.29B tokens from TCM sources, including web and books, supplemented with 5.63B tokens of general-purpose text to preserve the LLMs' foundational capabilities. As detailed in Appendix~\ref{ap_general-corpus}, incorporating general data does not impair TCM-specific learning, while effectively preserving the model's general reasoning abilities. In total, Stage 1 involves 11.92B tokens of text-only pretraining. To optimize training efficiency, we concatenate input samples into sequences of 4096 tokens, using the special token \texttt{<|endoftext|>} to delimit different documents. The model is trained to predict all tokens except the \texttt{<|endoftext|>} tokens. Pretraining is conducted for 1 epoch with a batch size of 256, a learning rate of 5e-5, and a warmup ratio of 0.005.

\paragraph{Stage 2: Multimodal Pre-training}  
We then pre-train on multimodal data to introduce TCM visual knowledge and basic audio understanding. This includes 1.17B tokens from TCM image-text data and 0.69B tokens from general image-text datasets (e.g., ShareGPT-4V), totaling 1.86B image-text tokens. For audio-text alignment, we include 0.03B tokens from TCM-specific audio-text data and 0.04B from general audio datasets to ensure basic audio comprehension. Additionally, we resample 1.75B tokens from the Stage 1 text corpus to maintain textual grounding during multimodal learning. Altogether, Stage 2 involves 3.84B tokens of multimodal pretraining. As in Stage 1, inputs are packed into 4096-token sequences to accelerate training. And only textual tokens are used for loss computation. We use a batch size of 128 and the same learning rate of 5e-5 for this stage.

\section{Data Leakage Prevention}

Preventing data leakage between training and evaluation is critical for reliable benchmarking. In this work, we adopted two strategies to mitigate leakage risk:

\subsubsection{Temporal Separation}

All training data were collected \textbf{before 2023}, while evaluation sets are from \textbf{late 2024 to 2025}, ensuring no overlap in time.

\begin{itemize}
    \item \textbf{Common Crawl}: 2017–2023 (no data from 2024 or later).
    \item \textbf{WeChat articles}: collected prior to June 2023.
    \item \textbf{TCM textbooks}: published no later than 2021.
\end{itemize}

In contrast, evaluation benchmarks are based on recent official exams. As shown in Table~\ref{tab:evaluation_dates}, all evaluation dates postdate the training period.

\begin{table}[H]
\centering
\caption{Evaluation benchmark sources and their exam dates. All were administered after training data collection.}
\resizebox{1.0\linewidth}{!}{
\begin{tabular}{lc}
\toprule
\textbf{Evaluation Benchmark} & \textbf{Exam Date} \\
\midrule
TCM Pharmacist & April 20, \textcolor{blue}{2024} \\
TCM Physician & June 15 -- November 10, \textcolor{blue}{2024} \\
TCM Assistant Physician & June 15 -- November 9, \textcolor{blue}{2024} \\
TCM Graduate Entrance (2024) & Dec 21, \textcolor{blue}{2023} -- Feb 30, \textcolor{blue}{2024} \\
TCM Graduate Entrance (2025) & Dec 20, \textcolor{blue}{2024} -- Feb 30, \textcolor{blue}{2025} \\
\bottomrule
\end{tabular}
}
\label{tab:evaluation_dates}
\end{table}

\subsubsection{Content-Based Filtering}

To further prevent leakage, we applied a strict content overlap filter. For the TCM-Vision Benchmark, we followed the protocol of Med-PaLM 2~\cite{medpalm2}, removing any evaluation instance with a 64-character or longer exact match with any training source, including textbooks and scraped corpora.

These two safeguards, temporal separation and content-based filtering, jointly ensure that our evaluation results reflect true generalization rather than memorization.

\section{Instruction Fine-tuning Dataset}
\label{ap_instruction_dataset}

\begin{figure}[ht!]
\centering
\begin{prompt}[title={The Prompt for Verifier}] 
{
{\textless}Model Response{\textgreater}  \newline
\textcolor{blue}{\texttt{\{Model Response\}}}\newline
\textless/Model Response\textgreater  \newline

{\textless}Reference Answer{\textgreater}  \newline
\textcolor{blue}{\texttt{\{Ground-true Answer\}}}\newline
{\textless}/Reference Answer{\textgreater}  \newline

You are provided with a model-generated response ({\textless}Model Response{\textgreater}) and a reference answer ({\textless}Reference Answer{\textgreater}). Compare the model response with the reference answer and determine its correctness. Your task is to simply output "True" if the response is correct, and "False" otherwise.
}
\end{prompt}
\caption{The prompt for the Deepseek-V3 verifier. \textcolor{blue}{\texttt{\{Model Response\}}} represents the output of the model to be verified. \textcolor{blue}{\texttt{\{Ground-true Answer\}}} represents the ground-truth answer or doctor response for TCM problems.}
\label{prompt-verifier}
\end{figure}

\paragraph{Text Instructions} We collected 92k authentic TCM-related questions from the Huatuo-26M doctor-patient dialogue corpus and the HuatuoGPT-o1 verifiable question set. These questions were then distilled using Deepseek-R1 to extract formal responses, omitting the reasoning components. Given that each question was paired with either a doctor's reply or a verifiable answer, we employed Deepseek-V3 to assess whether the responses passed verification, using the prompt illustrated in Figure \ref{prompt-verifier}. This process ultimately yielded a Chinese instructional dataset comprising 83k entries.

\begin{prompt}[title={The prompt for generating questions for TCM Vision Instructions}] 
{
 \textcolor{blue}{\texttt{\{image\}}}\newline

Please generate a Traditional Chinese medicine (TCM)-related question about the medical image I provide. The question should assess the model's visual capabilities. Avoid being too specific—design the question so that it requires the model to look at the image to answer. The question should demand strong visual understanding as well as some knowledge of Traditional Chinese medicine.
\newline

You may refer to the provided image caption and contextual information to improve the quality of the question. However, **do not mention or reference the caption or context in the question itself—assume they are not available**.
\newline

{\textless}caption{\textgreater}\newline
 \textcolor{blue}{\texttt{\{caption\}}}\newline
{\textless}/caption{\textgreater}\newline

{\textless}image context{\textgreater} ({\textless}image{\textgreater} represents the location of the image)\newline
 \textcolor{blue}{\texttt{\{image context\}}}\newline
{\textless}/image context{\textgreater}\newline

Please generate the question directly. Do not include any content unrelated to the task.
}
\end{prompt}
\captionof{figure}{\label{prompt-question}The prompt for generating questions for TCM Vision Instructions. Here, \textcolor{blue}{\texttt{{image}}} represents the TCM image. \textcolor{blue}{\texttt{{caption}}} and \textcolor{blue}{\texttt{{image context}}} represent the corresponding image caption and contextual information.}

\paragraph{Vision Instructions}  
We filtered TCM-related images with detailed captions from the books and web interleaved data. These were provided to GPT-4o, which first generated vision-related questions based on the image, caption, and context, using the prompt in Table \ref{prompt-question}. Then, GPT-4o was provided the image, caption, and context to generate the corresponding response using the prompt in Table \ref{prompt-answer}. This process yielded 65K high-quality visual instruction samples.

\begin{prompt}[title={The prompt for generating answers for TCM Vision Instructions}] 
{
 \textcolor{blue}{\texttt{\{image\}}}\newline

{\textless}question{\textgreater}\newline
 \textcolor{blue}{\texttt{\{caption\}}}\newline
{\textless}/question{\textgreater}\newline

You are now required to look at the image I provide and answer the user's question about Traditional Chinese Medicine ({\textless}question{\textgreater}). Make sure your response directly addresses the user's query, follows instructions well, and is as detailed and rich as possible, with the style and quality characteristic of GPT-4o.\newline

You may refer to the image caption and contextual information I secretly provide to you in order to improve the accuracy and completeness of your answer. However, **do not mention or reference the caption or context in your response—assume they are not available**.\newline

{\textless}caption{\textgreater}\newline
 \textcolor{blue}{\texttt{\{caption\}}}\newline
{\textless}/caption{\textgreater}\newline

{\textless}image context{\textgreater} ({\textless}image{\textgreater} represents the location of the image)\newline
 \textcolor{blue}{\texttt{\{image context\}}}\newline
{\textless}/image context{\textgreater}\newline

Please generate the answer directly. Do not include any unrelated content.
}
\end{prompt}
\captionof{figure}{\label{prompt-answer}The prompt for generating answers for TCM Vision Instructions. Here, \textcolor{blue}{\texttt{{image}}} represents the TCM image.  \textcolor{blue}{\texttt{{question}}} represents the generated visual question. \textcolor{blue}{\texttt{{caption}}} and \textcolor{blue}{\texttt{{image context}}} represent the corresponding image caption and contextual information.}

\paragraph{Audio Instructions}
The audio instruction set includes two parts. For speech instruction, we sampled 20K questions from the text instructions and converted them into speech using TTS. Each audio was paired with its original answer. Additionally, we added transcription-style prompts (e.g., “Please repeat exactly what was said”) to the same audio inputs, creating another 20K instruction pairs. For TCM-specific audio, such as cough {Placeholder} or heart sounds {Placeholder},  we converted these classification tasks into instruction format. For example: “Please listen to the patient's cough and determine if they have COVID-19. Answer: Yes or No.” 

\paragraph{Physiological Signal Instructions}  
For non-audio physiological signals, we convert all inputs into waveform format and prepend modality-specific tokens (e.g., "\texttt{<Smell>}", "\texttt{<Pulse>}", "\texttt{<ECG>}"). For smell, we use the {Placeholder} TCM smell dataset. For pulse, we collected 5k samples from pregnant and non-pregnant individuals (1:1 ratio), using 80\% for training. ECG data is sourced from {Placeholder} and {Placeholder} datasets. All classification tasks were reformulated into instruction-based prompts. For example: “Based on this pulse waveform, determine whether the person is pregnant. Answer ‘Yes' or ‘No.'”. To support extensibility, we provide a unified signal-processing pipeline that enables ShizhenGPT to adapt to a wide range of physiological modalities with minimal effort.

\section{Instruction Fine-tuning Details}
To complete the training pipeline, we merge all modality-specific instruction datasets (Table~\ref{tab:instruction-datasets}) and fine-tune the TCM-pretrained model jointly. All instructions are formatted using the Qwen2.5 prompt template, and only the output tokens corresponding to the model's response are included in the loss computation. The model is trained for 3 epochs with a learning rate of 5e-6, batch size of 128, and a warmup ratio of 4\%. This stage yields our final model, \textbf{ShizhenGPT}.

\section{Physiological Signal Datasets}
\label{ap_tcm_signal}

\paragraph{Pulse} Pulse signals capture fluctuations in cardiac rhythm and have historically played a vital role in assisting diagnosis within the framework of TCM.

\begin{itemize}
    \item \textbf{Pregnancy Identification Dataset}: In collaboration with a partner hospital, pulse signals were collected from 3,165 healthy women of childbearing age by licensed TCM practitioners with over two years of clinical experience. Data collection was conducted in a resting state using a pulse acquisition device, following the traditional “three positions and nine indicators” (Cun, Guan, Chi) method on both radial arteries. Each sample was annotated with gestational age, determined based on the last menstrual period or ultrasound findings. The final dataset was split into training and testing sets using an 80/20 ratio.
    \item \textbf{\texttt{AdBrc pulse}}~\cite{guo2022wrist} dataset: We adopted this publicly available dataset to enrich the pulse diagnosis task. It includes nine health conditions, each with 218 samples, with signals collected from the Guan position based on the TCM Cun-Guan-Chi theory. This version provides two types of pulse signals: continuous and cycled. We used only the continuous signals from the "continues" folder, which contain raw data acquired via pressure sensors. All data were converted into a unified time-series format and split into training and testing sets using an 80/20 ratio.
\end{itemize}

\paragraph{Smell} We used the \textbf{\texttt{CUHKSZ-Odors}}~\cite{guo2022wrist} dataset to evaluate the model's ability to interpret olfactory signals in TCM. The dataset, intended solely for research and requiring prior authorization via the official platform, must not be used commercially or shared with unauthorized parties. Collected from real patients and labeled by professional physicians, it covers eight categories: Health, Diabetes, Nephrosis, Bile Diseases, Liver Disease, Lung Disease, Gastritis, and Digestive Diseases. Each category contains 120 samples—100 clean and 20 with sampling interference. Each sample is a 1153×9 matrix, representing data from nine gas sensors with 1153 readings per sensor.

\paragraph{Sound} In TCM auscultation, patient voice characteristics are an important diagnostic cue. To assess our model's audio classification ability, we used the \textbf{\texttt{COUGHVID dataset}}~\cite{orlandic2021coughvid}, which contains over 25,000 crowdsourced cough recordings across diverse demographics and COVID-19 statuses. Of these, 2,800 samples were expert-labeled by physicians, forming one of the largest annotated cough datasets for respiratory condition detection. This resource supports robust model evaluation in sound-based diagnostic tasks. To address the severe class imbalance, we adjusted the dataset to achieve equal numbers of samples for both labels. An 80/20 split was then applied for training and testing.

\paragraph{Modern Medicine Modalities} Given the relatively small scale of TCM-specific modality data, we supplemented our experiments with comparable modalities from modern medicine to ensure comprehensive validation of our approach.

\begin{itemize}
    \item \textbf{Electrocardiogram (ECG) signals}, widely adopted in modern medicine, provide insights into cardiac rhythm and support clinical assessments of vital signs and cardiovascular conditions. In TCM, pulse diagnosis—based on palpating the patient's pulse—serves a comparable function, capturing variations in heartbeat rhythm as a form of time-series data. Leveraging \textbf{\texttt{PTB Diagnostic ECG Database}}~\cite{bousseljot1995nutzung} and \textbf{\texttt{MIT-BIH Arrhythmia Database}}~\cite{moody2001impact}, we validate the effectiveness of our proposed method on this analogous modality. We adopt the original training and testing split provided by the dataset for model training and evaluation.
    % \item \textbf{Heart sounds} capture physiological changes in the heart through acoustic signals and are commonly used for the detection of certain cardiovascular conditions. We utilize the dataset, \textbf{\texttt{The CirCor DigiScope Phonocardiogram Dataset}}~\cite{oliveira2022circor}, recordings as audio input to assess whether our method can effectively classify such acoustic data, thereby providing preliminary support for enabling LLM-based approaches to TCM auscultation (闻). The original dataset was divided into training and testing sets using an 80/20 split.
    \item \textbf{Heart sounds} capture physiological changes in the heart through acoustic signals and are commonly used for the detection of certain cardiovascular conditions. We utilize the dataset, \textbf{\texttt{The CirCor DigiScope Phonocardiogram Dataset}}~\cite{oliveira2022circor}, recordings as audio input to assess whether our method can effectively classify such acoustic data, thereby providing preliminary support for enabling LLM-based approaches to TCM auscultation (listening diagnosis). The original dataset was divided into training and testing sets using an 80/20 split.
\end{itemize}

\section{Benchmarks}
\label{ap_benchmark}
To comprehensively evaluate the TCM capabilities of our model, we design a comprehensive multimodal benchmark suite across all modalities: text, vision and physiological signals. The benchmarks are categorized as follows:

\begin{itemize}
\item \textbf{Text Benchmarks}  
We evaluate textual TCM knowledge using two sources: (1) official national exams and (2) public QA benchmarks. For the former, we collect the most recent national exams in China, including the 2024 National TCM Pharmacist Licensing Exam, the 2024 TCM Physician Licensing Exam, and the 2024 Assistant Medical Exam. All questions are less than a year old to ensure recency and minimize training overlap.

\item \textbf{Vision Benchmark}  
To assess visual understanding in TCM, we construct a benchmark from seven authoritative atlases—explicitly excluded from our training set. Each book contributes to a specific task:  
\textit{Chinese Herbal Medicine Color Atlas} (herb recognition),  
\textit{Chinese Medicinal Materials} (TCM material classification),  
\textit{Practical Atlas of Decoction Pieces} (decoction recognition),  
\textit{Holographic Inspection Atlas} (holistic diagnosis),  
\textit{Tongue Diagnosis Atlas} (tongue inspection),  
\textit{Eye Diagnosis Atlas} (eye pattern recognition),  
\textit{Palm Symptom Atlas} (palm inspection), and  
\textit{Tuina Techniques Atlas} (massage gesture recognition).
All visual questions follow a multiple-choice format (one image, one question, four options, one correct). For images with explicit captions (e.g., herbs), we randomly select three nearby captions as distractors and use the correct one as the answer. For more complex content (e.g., Tuina), we prompt GPT-4o with the image, its caption, and local context to generate corresponding questions. The final vision benchmark includes 7,204 questions across seven TCM-specific visual domains.

\item \textbf{Human Benchmark} Details at Appendix \ref{app:humaneval}.

\item \textbf{Signal Benchmark}  For signal-based modalities such as smell, auscultation, and ECG, we evaluate on several publicly available datasets:  
COUGHVID~\cite{orlandic2021coughvid} for respiratory sounds and cough,  
The CirCor DigiScope Phonocardiogram Dataset~\cite{oliveira2022circor} for heart sounds,  
AdBrc pulse~\cite{guo2022wrist} for smell signals,  
PTB Diagnostic ECG Database~\cite{bousseljot1995nutzung} and MIT-BIH Arrhythmia Database~\cite{moody2001impact} for ECG signals.  
We use their official test splits for evaluation. For pulse wave analysis, we construct a custom benchmark for pregnancy detection based on pulse signals, holding out 20\% of the data as the test set.

\end{itemize}

\section{Human Evaluation}
\label{app:humaneval}

We selected 6 powerful LLMs to compare against ShizhenGPT, using a set of 90 real-world TCM cases from \textit{Journal of Chinese Clinical Case Achievements Database}. ShizhenGPT's responses were compared pairwise with those of each model. After randomly shuffling the model order, we invited three licensed TCM practitioners to evaluate the responses based on correctness, completeness, and clinical insight. Specifically, each doctor was tasked with assessing 180 pairs of model answers and received a compensation of 300 RMB. For each pair, they were asked to choose the response that better aligns with TCM medical standards, or mark them as equally aligned if no clear preference could be made (see Figure~\ref{humaneval-prompt} for evaluation criteria).

\begin{prompt}[title={The prompt for human evaluation}] 
Please evaluate the responses of two language models to the same TCM-related question and indicate which answer you professionally prefer. Your judgment should be based primarily on the accuracy of TCM theory, the proper use of professional terminology, the completeness of the response, and its clinical relevance.\newline

If one response is clearly more rigorous, reasonable, or clinically valuable—or if the other contains obvious errors, unclear expression, or lacks professionalism—please label it as “Model 1” or “Model 2” accordingly. If both responses are of similar quality and no clear preference can be made, please mark it as “Tie”.\newline

During the evaluation, please avoid being influenced by writing style, expression, or response length. Focus solely on the accuracy and practical value of the TCM content. We sincerely appreciate your professional judgment and support!

\end{prompt}
\captionof{figure}{\label{humaneval-prompt}The instruction given to the doctors for human evaluation.}

\section{Evaluation Protocol}

For the Traditional Chinese Medicine (TCM) evaluation, we use the average score of five assessments as the final result. In contrast, for the visual evaluation, we only use the result of a single assessment due to the large size of the evaluation task. All experiments are conducted under a zero-shot setting. A random seed of 42 is used for all training tasks to ensure the reproducibility of the model.

\section{Limitations and Potential Risks}

\paragraph{Limitations}
This work presents several potential limitations:
\squishlist
\item \textbf{Insufficient TCM Signal Data} Despite the large-scale collection of text and image data, high-quality signal data (e.g., smell, pulse) remains scarce. Most of it comes from limited public datasets, which constrains the model’s ability to fully develop and generalize in these modalities.
\item \textbf{Lack of Real-World Clinical Testing} Although expert evaluations have been conducted, ShizhenGPT has not been tested in actual clinical settings. The absence of patient-level feedback limits our understanding of its practical effectiveness and safety.
\item \textbf{Incomplete Modal Coverage} The current modalities include vision, sound, and selected physiological signals, but other modalities such as tactile sensation are still missing.
\squishend

Despite these limitations, we hope this work will inspire deeper collaboration between computer scientists and medical practitioners, particularly in building richer multimodal datasets to advance AI in Traditional Chinese Medicine.

\paragraph{Potential Risks}
As a multimodal large language model (LLM) for Traditional Chinese Medicine, the current version of ShizhenGPT may still generate content that includes hallucinations or factual inaccuracies. Therefore, it is intended solely for scientific research purposes and is not suitable for deployment in real-world applications. To mitigate potential risks, we impose strict limitations on its usage: the model must not be applied in clinical practice or any industry-related scenarios where inaccuracies could result in adverse consequences. We strongly emphasize the ethical responsibility of users to comply with these restrictions to ensure the safety and integrity of downstream applications.

\begin{figure*}[ht!]
    \centering
    \includegraphics[width=0.9\linewidth]{{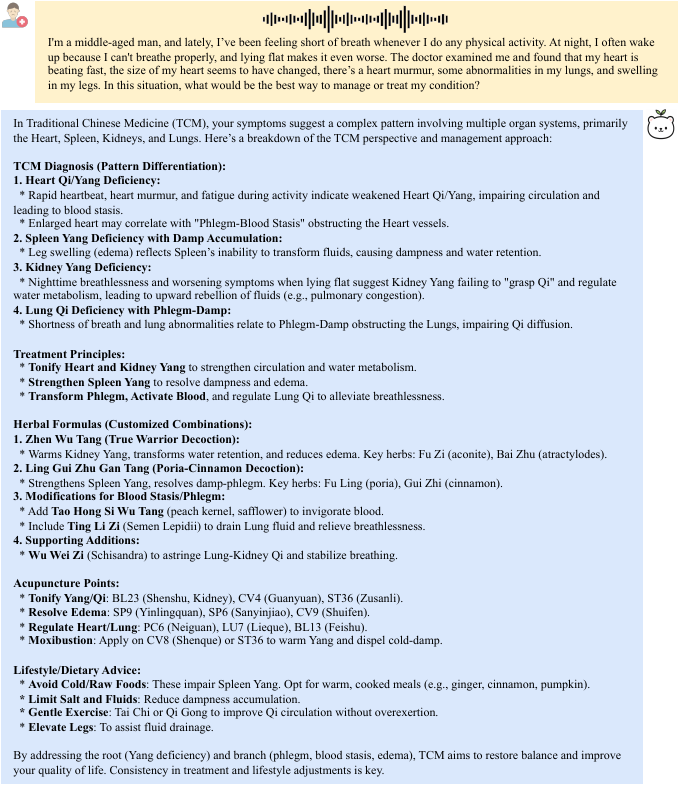}}
    \caption{An example of ShizhenGPT with audio input}
    \label{app-case2}
\end{figure*}

\begin{figure*}[ht!]
    \centering
    \includegraphics[width=0.9\linewidth]{{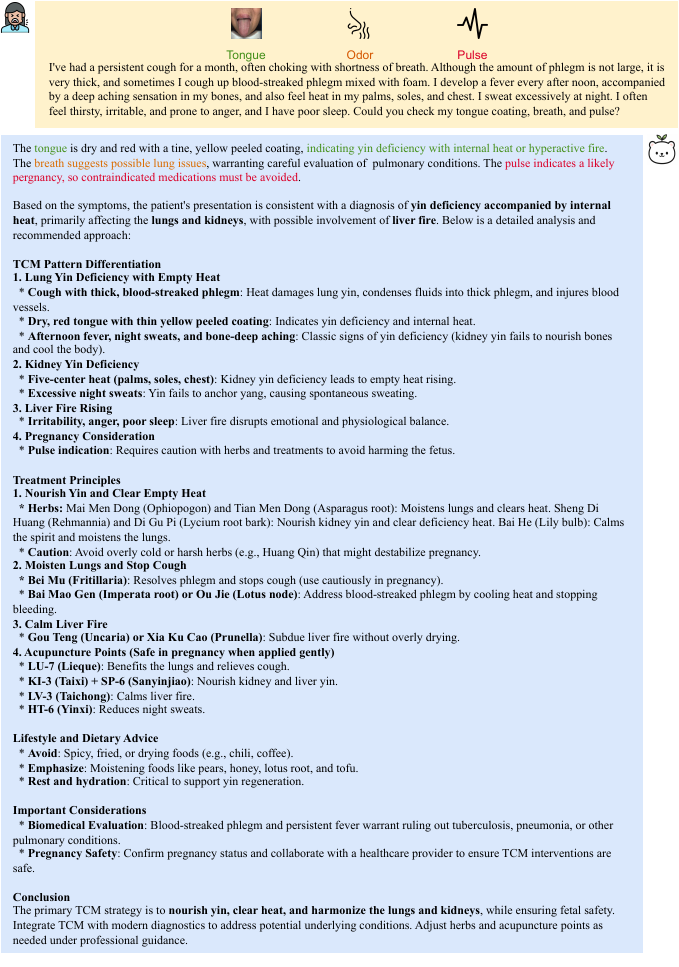}}
    \caption{An example of ShizhenGPT with multimodal inputs}
    \label{app-case1}
\end{figure*}

\begin{figure*}[ht!]
    \centering
    \includegraphics[width=0.9\linewidth]{{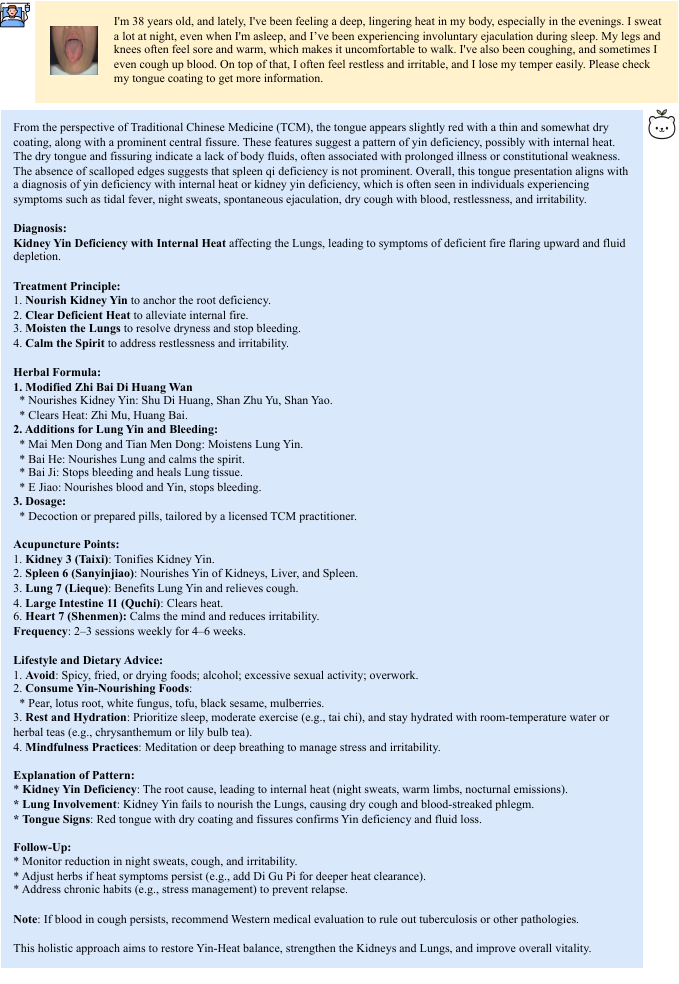}}
    \caption{An example of ShizhenGPT with image input}
    \label{app-case3}
\end{figure*}

\section{Output Examples}
\label{app:case}
Several real output examples of ShizhenGPT are presented here for reference. Figure \ref{app-case1} illustrates the performance of our model when processing inputs from multiple modalities simultaneously. The model effectively distinguishes and integrates information from various sources into its diagnostic reasoning, demonstrating the potential of multimodal fusion to support TCM consultations. Figure \ref{app-case2} showcases an example of patient interaction through direct voice input, where the patient verbally describes symptoms and the model generates diagnostic reasoning, enhancing the system's interactivity. Lastly, Figure \ref{app-case3} presents the model's diagnostic results based on tongue coating analysis, highlighting its capability in visual perception and understanding.

\end{document}